\documentclass[a4paper,fleqn]{cas-dc}

\ExplSyntaxOn
\cs_gset:Npn \__first_footerline:
  { \group_begin: \small \sffamily \__short_authors: \group_end: }
\ExplSyntaxOff

\usepackage[numbers,sort&compress]{natbib}
\usepackage{amsmath,amssymb}
\usepackage{multirow}
\usepackage{booktabs}
\usepackage{tabularx}
\usepackage{graphicx}
\usepackage{placeins}
\usepackage{afterpage}
\usepackage{xspace}
\usepackage{xcolor}
\usepackage[capitalise]{cleveref}
\crefname{figure}{\figurename}{\figurename}
\Crefname{figure}{\figurename}{\figurename}
\usepackage{hyphenat}

\DeclareMathOperator{\diag}{diag}
\newcommand{\KCHDAT}{MEN\xspace}

\usepackage{pifont}
\newcommand{\cmark}{\ding{51}}
\newcommand{\xmark}{\ding{55}}

\begin{document}
\let\WriteBookmarks\relax
\def\floatpagepagefraction{1}
\def\textpagefraction{.001}

\shorttitle{Flexible Multi-Stack Super-Resolution of Clinical Brain MRI Using Hetero-Orientation Variational Encoding: Expert Reader Evaluation in a Nigerian Cohort}
\shortauthors{Muhammad et~al.}

\title[mode=title]{Flexible Multi-Stack Super-Resolution of Clinical Brain MRI Using Hetero-Orientation Variational Encoding: Expert Reader Evaluation in a Nigerian Cohort}

\author[1,3]{Kabir Hamzah Muhammad}[orcid=0000-0002-8459-383X]\ead{kabir.muhammad@kcl.ac.uk}\cormark[1]
\credit{Conceptualization, Methodology, Software, Validation,
Formal analysis, Investigation, Visualization,
Writing -- original draft, Writing -- review \& editing}

\author[1]{Lorena {Garcia-Foncillas Macias}}[orcid=0000-0003-1015-4890]\ead{lorena.garcia-foncillas_macias@kcl.ac.uk}
\credit{Data curation}

\author[1]{Soumya Snigdha Kundu}[orcid=0000-0002-4130-7725]\ead{soumya_snigdha.kundu@kcl.ac.uk}
\credit{Resources}

\author[5]{Emily Guilhem}[orcid=0000-0001-9727-1090]\ead{emily.guilhem6@nhs.net}
\credit{Investigation, Validation}

\author[1,12]{Jon Cleary}[orcid=0000-0001-8836-9934]\ead{jon.cleary@kcl.ac.uk}
\credit{Investigation, Validation}

\author[7]{Amal Saleh Nour}[orcid=0000-0001-8922-0031]\ead{amal.saleh@aau.edu.et}
\credit{Investigation, Validation}

\author[8]{Yodit Abraham Yaynishet}[orcid=0009-0003-6153-7952]\ead{yaynishety@chop.edu}
\credit{Investigation, Validation}

\author[9]{Chinasa Kalaiwo}[orcid=0000-0002-7325-1607]\ead{chinasakalaiwo@gmail.com}
\credit{Investigation, Validation}

\author[10]{Peter Chibuta}[orcid=0009-0002-0129-6790]\ead{peter.chibuta@moh.gov.zm}
\credit{Investigation, Validation}

\author[11]{Ugumba Kwikima}[orcid=0009-0001-5229-531X]\ead{ukwikima@gmail.com}
\credit{Investigation, Validation}

\author[2,3]{Udunna Anazodo}\ead{udunna.anazodo@mcgill.ca}
\credit{Writing -- review \& editing}

\author[1,4]{Jonathan Shapey}\ead{jonathan.shapey@kcl.ac.uk}
\credit{Data curation, Methodology, Funding acquisition,
Investigation, Resources, Supervision}

\author[1]{Tom Vercauteren}\ead{tom.vercauteren@kcl.ac.uk}\cormark[1]
\credit{Conceptualization, Funding acquisition, Investigation,
Methodology, Project administration, Resources, Supervision,
Writing -- review \& editing}

\affiliation[1]{organization={School of Biomedical Engineering \& Imaging Sciences, King's College London},
            city={London},
            country={United Kingdom}}

\affiliation[2]{organization={Montreal Neurological Institute, McGill University},
            city={Montreal},
            country={Canada}}

\affiliation[3]{organization={Medical Artificial Intelligence, Crestview Radiology Ltd.},
            city={Lagos},
            country={Nigeria}}

\affiliation[4]{organization={Department of Neurosurgery, King's College Hospital NHS Foundation Trust},
            city={London},
            country={United Kingdom}}

\affiliation[5]{organization={King's College Hospital NHS Foundation Trust}, 
            city={London},
            country={United Kingdom}}

\affiliation[7]{organization={School of Medicine, College of Health Sciences, Addis Ababa University}, 
            city={Addis Ababa}, 
            country={Ethiopia}}

\affiliation[8]{organization={Children's Hospital of Philadelphia}, 
            city={Philadelphia}, 
            country={United States}}

\affiliation[9]{organization={African Medical Centre of Excellence}, 
            city={Abuja}, 
            country={Nigeria}}

\affiliation[10]{organization={University Teaching Hospital}, 
            city={Lusaka}, 
            country={Zambia}}

\affiliation[11]{organization={Muhimbili Orthopedic Institute \& Muhimbili University of Health and Allied Sciences}, 
        city={Dar Es Salaam}, 
        country={Tanzania}}

\affiliation[12]{organization={Department of Radiology, Guy's and St. Thomas' NHS Foundation Trust}, 
        city={London},
        country={United Kingdom}}

\cortext[1]{Corresponding author.}

\begin{abstract}
High-resolution isotropic brain MRI is important for neuroradiological assessment, yet routine clinical protocols often rely on fast, anisotropic 2D acquisitions, particularly in resource-constrained settings. These acquisitions may differ in orientation, alignment and anatomical coverage, resulting in partially overlapping image stacks. Existing learning-based super-resolution methods typically rely on single, under-constrained inputs or fixed multi-stack configurations with idealised acquisition geometry.
We introduce MS-HVED, a sequence-agnostic framework for flexible multi-stack brain MRI super-resolution. Available axial, coronal and sagittal stacks are independently encoded into multi-scale Gaussian posteriors and fused using a field-of-view-aware Product-of-Gaussians formulation that limits contributions from unsupported regions. Physics-informed domain randomisation simulates heterogeneous clinical acquisition conditions, including obliqueness, partial coverage and imaging artefacts. We evaluate dataset-specific and sequence-agnostic models trained on T1- and T2-weighted healthy and clinical brain MRI.
MS-HVED achieved competitive or superior reconstruction fidelity across synthetic benchmarks and improved downstream tissue segmentation relative to low-resolution inputs. To assess generalisation beyond the training cohorts, we performed external clinical validation on 30 routine brain MRI examinations from a Nigerian cohort, representing an under-represented population in medical imaging research. These examinations formed the largest component of a blinded reader study involving 44 real cases and seven expert neuroradiologists. In the Nigerian cohort, MS-HVED was preferred over the supervised baseline for overall image quality in 187 of 210 paired assessments. Across all clinical cohorts, MS-HVED achieved higher overall-quality scores with all available inputs ($5.93$ vs.\ $4.49$, $p<.001$), with the preference persisting when axial or coronal inputs were omitted ($p<.001$).
These findings demonstrate the potential of flexible, acquisition-aware reconstruction to improve the visual quality of heterogeneous routine brain MRI, including examinations from an under-represented African clinical population. Code and trained weights are available at \url{https://github.com/cai4cai/ms-hved}.
\end{abstract}


\maketitle

\section{Introduction}
\label{sec:intro}

High-resolution isotropic brain MRI is important for accurate diagnosis, surgical planning, and longitudinal monitoring of neurological disease~\cite{vanreeth2012srmri,li2022clinicalsr}. In routine clinical practice, and particularly in resource-constrained settings where scanner availability, patient throughput, and acquisition time are major constraints, brain MR examinations are often conducted with fast 2D multi-slice protocols with thick slices, employed to keep scan time low, mitigate motion, and maintain signal-to-noise ratio (SNR)~\cite{vanreeth2012srmri,li2022clinicalsr,acr2026headneck}. Although these protocols achieve high in-plane resolution, they suffer from poor through-plane resolution, producing anisotropic 3D stacks that hinder volumetric analysis and multi-planar review. To partially compensate for this limitation, clinical protocols commonly acquire multiple stacks across axial, coronal and sagittal orientations~\cite{acr2026headneck}.

Super-resolution reconstruction (SRR) seeks to recover a single isotropic high-resolution (HR) volume from one or more anisotropic low-resolution (LR) stacks. Multi-stack SRR exploits complementary content of orthogonal acquisitions, where each stack provides high-fidelity information in its in-plane directions while lacking the through-plane detail~\cite{vanreeth2012srmri,gholipour2010robust}. The central methodological challenge is therefore to fuse these incomplete, orientation-specific observations into a geometrically consistent volumetric representation, ideally without per-subject optimisation or expensive iterative slice-to-volume registration.

The classical line of work formulates SRR as a model-based inverse problem solved by iterative slice-to-volume registration and optimisation~\cite{gholipour2010robust,NiftyMiC,SVoRT,NeSVoR}. These methods are well established in fetal MRI, where extreme inter-slice motion makes registration a dominant difficulty. However, they are computationally expensive and can be brittle in heterogeneous adult cohorts and low-resource clinical settings, where fast 2D acquisitions remain the norm. Learning-based SRR methods can be broadly divided into supervised and self-supervised methods. Supervised models are trained once on paired LR-HR datasets and then applied directly to new scans~\cite{srcnn,Huang2025}. However, many learned SRR methods operate on a single anisotropic stack, making the reconstruction problem severely under-constrained in the through-plane direction and potentially prone to blurring, loss of fine structure, or hallucinated anatomy~\cite{synthsr,inversesr,mriflow2023}.

Beyond the through-plane ambiguity, single-stack reconstruction is also limited by the anatomical extent of the selected acquisition. In routine clinical protocols, axial, coronal and sagittal stacks may be prescribed with different partial fields-of-view (FOVs)~\cite{acr2026headneck,lei2023fov}. Consequently, anatomy or pathology that is absent from one orientation may still be visible in another. For example, an axial stack prescribed with limited superior-inferior coverage may exclude an extra-axial lesion near the vertex, whereas the same lesion may be captured in the sagittal stack. This type of orientation-specific variation in anatomical coverage is shown in the appendix (\Cref{fig:partial_fov}). A method that reconstructs from the axial stack alone cannot recover such clinically relevant information, regardless of its super-resolution capacity. This motivates multi-stack SRR methods that exploit complementary orientation-specific coverage rather than treating each stack in isolation.

Self-supervised learned reconstruction methods such as SMORE~\cite{smore}, ECLARE~\cite{eclare} and TripleSR~\cite{triplesr} reduce the need for paired supervised training data by optimising a model using only the available anisotropic scan. However, these approaches typically require a computationally heavy per-subject optimisation at inference and therefore do not provide a single reusable model that can immediately reconstruct arbitrary subsets of axial, coronal and sagittal inputs. Flexible fusion of missing or incomplete inputs has been studied in multi-modal segmentation, including HeMIS~\cite{hemis} and U-HVED~\cite{uhved}, but analogous flexibility remains less explored for multi-stack MRI super-resolution, where the absence arises not only from absent input orientations but also from spatially varying FOV within each orientation.

\begin{figure*}[pos=t]
    \centering
    \includegraphics[width=\textwidth]{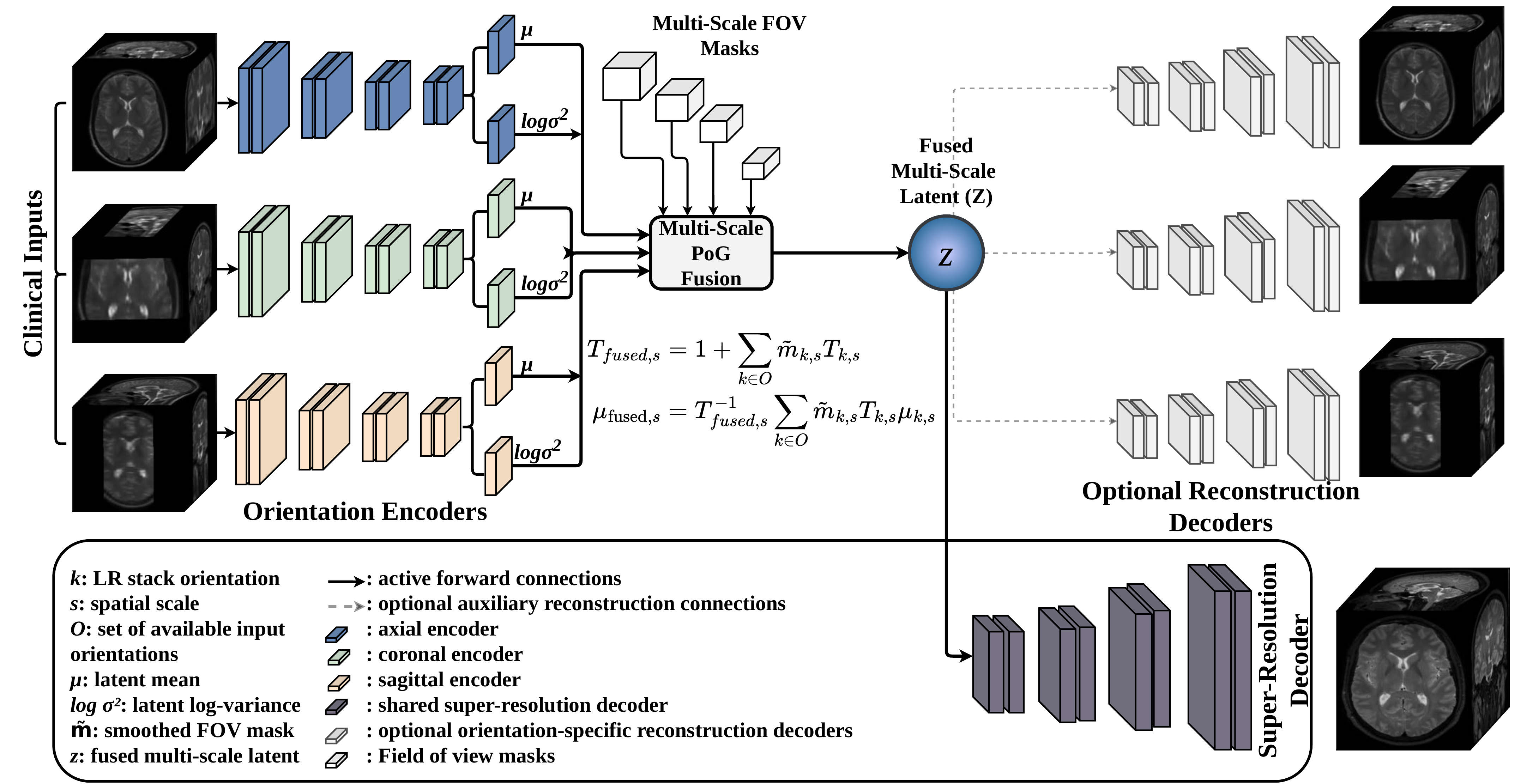}
    \caption{\textbf{Overview of the proposed MS-HVED framework.} 
Given any non-empty subset of axial, coronal and sagittal anisotropic stacks, each available orientation is processed by an independent variational encoder that predicts latent multi-scale Gaussian posterior parameters. The resulting orientation-specific latent posteriors are fused voxel-wise using a Multi-Scale Product-of-Gaussians, where per-orientation FOV masks modulate each stack's precision contribution according to its spatial extent. Samples from the fused latent multi-scale posterior are passed through a shared super-resolution decoder to reconstruct the isotropic HR volume. Optional orientation-specific decoders reconstruct the input stacks.}
    \label{fig:framework}
\end{figure*}

We introduce \emph{MS-HVED}, a flexible hetero-orientation variational framework for multi-stack brain MRI super-resolution. As shown in \Cref{fig:framework}, each available orientation is encoded independently into a latent multi-scale Gaussian posterior, and the available posteriors are fused using a Product-of-Gaussians (PoG). Because fusion is defined over the set of available orientations, the same network can reconstruct from any non-empty subset $\mathcal{O}\subseteq\{\text{Ax},\text{Cor},\text{Sag}\}$ at inference. This allows the model to use one, two or three input stacks without retraining or changing the network architecture.

This flexible fusion is paired with explicit modelling of the acquisition geometry encountered in real clinical multi-stack scans. Axial, coronal and sagittal stacks are rarely perfectly aligned with a canonical reference frame. Each stack may be prescribed at an oblique angle, and after registration to a common isotropic grid the resampled stack can occupy a rotated and partially valid grid, with invalid or weakly supported regions near the volume boundaries. A simple preprocessing workaround is to trim poorly supported edge slices after resampling, but this is a heuristic outside the reconstruction model and often discards useful measurements. MS-HVED instead makes stack prescription part of both training and inference, specifically, the forward model simulates acquisition obliqueness and partial-FOV, and the fusion module receives per-orientation FOV masks that modulate each stack's precision contribution at the voxel level.

Evaluating multi-stack MRI super-resolution is also challenging. Synthetic LR-HR experiments provide known references and controlled acquisition geometry, but may not fully capture the mismatch observed in real clinical data. Paired real LR-HR acquisitions provide a more realistic reference, but are scarce and when available, the LR and HR scans are acquired separately and may differ in contrast, intensity scale, motion state, etc. Finally, many clinically important datasets have no matched isotropic HR reference, requiring qualitative, expert-reader or task-based evaluation. We therefore evaluate MS-HVED using complementary protocols: full-reference quantitative metrics on synthetic benchmarks, blinded expert-reader evaluation on reference-free clinical cohorts as well as downstream tissue-segmentation analysis.

Across multiple experiments, MS-HVED achieves best or competitive reconstruction quality while retaining flexibility across input configurations. Ablation experiments show that oblique partial-FOV simulation and FOV-aware fusion are important for robust reconstruction under realistic acquisition geometry. On paired real healthy-control data, MS-HVED improves perceptual quality and structural similarity over the baselines. In blinded expert review, MS-HVED received higher ratings than the strongest baseline for overall quality, SNR, artefact reduction, and anatomical consistency, and this preference persisted when specific input stacks were unavailable.

The main contributions of this work are:
\begin{enumerate}
    \item \textbf{Flexible hetero-orientation variational fusion.} We formulate multi-stack brain MRI SRR as a hetero-orientation fusion problem, where each available orientation contributes a multi-scale Gaussian posterior to a shared latent representation. A single model can reconstruct from any non-empty subset of axial, coronal and sagittal inputs without retraining.

    \item \textbf{FOV-aware Product-of-Gaussians fusion.} We extend PoG fusion with per-orientation, per-voxel FOV masks that encode the spatial support of each stack. Unsupported regions purposely do not contribute to the fused posterior, while smooth mask boundaries reduce discontinuities near stack edges.

    \item \textbf{Physics-informed domain randomisation for clinical multi-stack MRI.} We propose a synthetic forward model that simulates anisotropic resolution, slice-profile effects, oblique stack prescription, partial-FOV coverage, motion ghosting, k-space artefacts, bias fields and Rician noise, enabling supervised training without requiring large paired LR-HR clinical datasets.

    \item \textbf{Spectrally normalised variational architecture.} We avoid activation normalisation layers in the reconstruction backbone to preserve intensity relationships between the inputs and reconstructions, and stabilise posterior prediction using spectral normalisation.

    \item \textbf{Comprehensive evaluation across synthetic, real (paired and reference-free) clinical settings.} We evaluate reconstruction fidelity, missing-orientation robustness, sequence-agnostic generalisation, downstream segmentation utility and clinical image quality, providing a comprehensive assessment than voxel-wise similarity metrics alone.

    \item \textbf{Open-source implementation.} Our code and model weights are open-source at \url{https://github.com/cai4cai/ms-hved} to support reproducibility and future benchmarking.
\end{enumerate}

\section{Related work}
\label{sec:related}

\subsection{Single-stack MRI super-resolution}
Convolutional super-resolution networks were introduced for natural images by~\cite{srcnn} and have since been transferred to MRI. 
\citet{pham2019cnn} apply 3D CNNs to recover isotropic resolution from a single thick-slice acquisition. SMORE~\cite{smore} achieves anti-aliasing and through-plane super-resolution from a single anisotropic volume in a self-supervised framework by exploiting the higher resolution of in-plane patches. ECLARE~\citep{eclare} extends this idea by estimating the slice excitation profile via ESPRESO~\citep{espreso} and training a per-subject network on cross-planar in-plane patches. More recent arbitrary-scale approaches use implicit or explicit neural representations to handle variable slice spacing~\cite{wang2024sainr,pang2024nexpr}, and diffusion models have been adapted to brain and body MRI SR with mechanisms to reduce sampling cost~\cite{han2024arbitrarydiffusion,zhao2025partialdiffusion}.

While powerful, single-stack methods remain intrinsically limited by the information available in the selected acquisition~\cite{vanreeth2012srmri,antun2020instabilities,bhadra2021hallucinations}. First, through-plane detail is under-constrained when only one anisotropic view is observed. Missing high-frequency structure must be inferred from learned priors or within-stack context, which can lead to blurring, hallucinated anatomy, or contrast inversions on out-of-distribution clinical data~\cite{antun2020instabilities,bhadra2021hallucinations}. Second, and more importantly for routine multi-orientation protocols, a single stack is limited by its anatomical extent. If a pathology or structure lies outside the prescribed FOV of the selected stack, it cannot be recovered by any single-stack SR method. 

A complementary line of work pre-trains generative priors and uses them at inference. SynthSR~\cite{synthsr} demonstrates that aggressive domain randomisation, i.e. sampling contrasts, resolutions, and artefacts from vast distributions during training, enables strong cross-domain generalisation in a single feed-forward pass, but the model still operates on a single stack at inference. InverseSR~\cite{inversesr} formulates SR as latent-diffusion inversion using a 3D brain-MRI prior, and MRIFlow~\cite{mriflow2023} uses a normalising flow with a frequency prior. These methods produce sharp reconstructions, but they require iterative inference or a learned prior over a target distribution and do not provide a natural mechanism for fusing complementary information from multiple anisotropic stacks.

\subsection{Multi-stack MRI super-resolution}
Multi-stack super-resolution and slice-to-volume reconstruction (SVR) have been studied extensively in fetal MRI, where extreme inter-slice motion makes registration the dominant difficulty. Classical optimisation based pipelines~\cite{gholipour2010robust,NiftyMiC} explicitly model the acquisition forward problem and alternate between volume reconstruction and slice-to-volume registration. They produce high-quality reconstructions but are computationally expensive and sensitive to registration failures, limiting their scalability to large heterogeneous clinical cohorts. Learning-based variants speed up the registration step using transformers~\cite{SVoRT}, replace the volume parameterisation with implicit neural representations~\cite{NeSVoR}, or learn slice-profile conditioned generative networks~\cite{sptsr2023}. Most recently, \cite{Huang2025} learn multi-view interpolation weights for fetal-brain SRR.

In adult brain MRI, where inter-slice motion is typically more moderate, several works specifically address learned multi-stack reconstruction. TripleSR~\cite{triplesr} fuses two orthogonal anisotropic stacks using an implicit neural representation trained with a sparse coordinate-based loss in continuous space. It is self-supervised and effective in its two-view setting, but it performs computationally heavy per-subject optimisation and does not directly support a third stack or a missing input orientation within one reusable model. Recent supervised orthogonal-scan methods also show the value of combining complementary anatomical views with 3D CNNs~\cite{tian2024orthogonal}, but they still assume a fixed input set. This motivates evaluating fixed-input baselines alongside a flexible model that can operate on arbitrary subsets of axial, coronal and sagittal stacks.

Most learning-based multi-stack SRR studies simplify either the input availability problem or the acquisition geometry. In the first case, models are trained for a fixed set of orientations and cannot directly adapt when one orientation is absent. In the second case, stacks are often treated as axis-aligned fully overlapping volumes. However, real clinical stacks may be obliquely prescribed and may provide only partial FOV.

\subsection{Hetero-modal and probabilistic fusion}

The idea of fusing variable subsets of inputs through a shared probabilistic latent space comes from multi-modal segmentation. HeMIS~\cite{hemis} aggregates per-modality features using first and second-order statistics, supporting any non-empty subset of input modalities. U-HVED~\cite{uhved} reformulates this idea as a hetero-modal variational encoder-decoder where each modality contributes a Gaussian posterior to a shared latent space through Product-of-Experts fusion, and a single trained model performs joint segmentation and modality completion under arbitrary subsets of available modalities. 

Although these hetero-modal approaches provide a useful mechanism for handling missing input channels, they were developed primarily for segmentation problems rather than geometric super-resolution. In those setups, missingness is treated at the input level where a modality is either available or absent. Multi-stack SRR introduces an additional spatial form of missingness because of the anatomical extent of the inputs. 

\subsection{Domain randomisation}

A major obstacle to supervised SRR is the scarcity of paired LR-HR clinical data. \emph{Domain randomisation}, popularized in robotics~\cite{tobin2017domainrandomization} and adapted for medical imaging by SynthSR~\cite{synthsr} and SynthSeg~\cite{SynthSeg}, addresses this problem by training on synthetic data generated from broad distributions over contrast, resolution, and artefacts, on the premise that a sufficiently randomized training distribution covers the true clinical distribution as a special case. 

In MRI super-resolution, domain randomisation has most often focused on intensity contrast, resolution, noise, bias fields, and other appearance level or degradation level factors. These components are important for cross dataset robustness, but they do not fully capture the geometry of routine multi-stack acquisitions. Existing randomisation strategies therefore leave a gap between synthetic LR generation and the spatial patterns observed in clinical multi-orientation protocols. This motivates randomisation schemes that model not only image degradation, but also acquisition geometry and orientation-specific FOV.

\subsection{Evaluation and benchmarking of MRI super-resolution}

Benchmarking MRI super-resolution is challenging because reconstruction quality is not fully captured by a single metric. Full-reference measures such as PSNR, SSIM, MAE/RMSE and normalised cross-correlation are useful when a HR reference is available, and remain widely used in MRI reconstruction, synthesis and super-resolution studies. However, voxel-wise and structural metrics can favour overly smooth reconstructions, are sensitive to intensity normalisation and spatial misregistration, and may not reflect the preservation of clinically relevant anatomy~\cite{dohmen2025similarity,adamson2025deepfeature}. Perceptual or feature-based metrics provide complementary information about sharpness and structural plausibility, but they are often derived from natural-image feature spaces and should be interpreted cautiously in medical imaging~\cite{zhang2018lpips,adamson2025deepfeature}. In addition, high image-level similarity does not necessarily imply improved clinical or downstream utility, such as tissue segmentation, lesion assessment or diagnostic confidence~\cite{zhang2021binarysr,amoros2025biomedicalsr}.

Most MRI SR studies therefore use a combination of full-reference quantitative metrics and qualitative assessment. Early and recent learning-based SR methods commonly report PSNR, SSIM, RMSE/MAE or related measures on synthetically downsampled data, where an HR target is known by construction~\cite{pham2019cnn,smore,eclare,triplesr}. Other studies evaluate on paired LR-HR acquisitions~\cite{Ebner2019}, which better reflect real imaging conditions but introduce a pseudo-reference problem: the HR scan is acquired separately and may differ in contrast or noise. In such cases, quantitative scores measure similarity to the available HR acquisition rather than absolute reconstruction accuracy.

Several MRI SR and accelerated reconstruction studies have therefore incorporated expert reader studies to assess clinical image quality directly. ~\citet{chaudhari2018superresolution} evaluated musculoskeletal MRI SR using both quantitative metrics and a radiologist reader study, with readers assessing sharpness, contrast, artefacts, SNR and overall diagnostic quality. ~\citet{Ebner2019} evaluated upper-abdominal MR super-resolution reconstruction with expert readers, reporting subjective preference and anatomical clarity scores for clinically relevant structures. In brain MRI, ~\citet{rudie2022clinical} prospectively evaluated deep learning-based resolution enhancement using four board-certified neuroradiologists who scored SNR, anatomic conspicuity, overall image quality, artefacts and diagnostic confidence on a Likert scale. These studies show that expert review is often needed when the relevant endpoint is diagnostic interpretability rather than numerical agreement with a reference image.

Task-based evaluation provides another complementary route. In medical image SR, ~\citet{zhang2021binarysr} showed that improvements in conventional image-quality metrics do not necessarily translate into improved signal-detection, motivating evaluation against clinically or analytically meaningful tasks. More recent biomedical SR benchmarks have similarly assessed whether SR improves downstream segmentation or classification rather than only image similarity~\cite{amoros2025biomedicalsr}. Together, these works suggest that a robust SR evaluation should combine full-reference metrics where valid references exist, qualitative or expert assessment where references are unavailable or imperfect, and downstream task evaluation where the reconstructed images are intended to support automated analysis.
\section{Method}
\label{sec:method}

We first formulate multi-stack super-resolution under realistic clinical acquisition geometry (\Cref{sec:method:formulation}), then describe the MS-HVED architecture (\Cref{sec:method:arch}). The following sections introduce the spectrally normalised variational backbone (\Cref{sec:method:spectral}), FOV-aware Product-of-Gaussians fusion (\Cref{sec:method:fov}), and the physics-informed domain randomisation  (\Cref{sec:method:dr}). We then define the training objective (\Cref{sec:method:loss}) and inference procedure for clinical stacks (\Cref{sec:method:inference}).

\subsection{Problem formulation}
\label{sec:method:formulation}

Let $X \in \mathbb{R}^{H \times W \times D}$ denote the unknown isotropic high-resolution (HR) volume on a canonical/reference grid. We observe a set of anisotropic low-resolution stacks $\mathcal{Y} = \{Y_k\}_{k \in \mathcal{O}}$ where $\mathcal{O} \subseteq \{\text{Ax}, \text{Cor}, \text{Sag}\}$ is the available subset of orientations. 
For compatibility with standard neural network architectures, all low-resolution stacks are resampled to the target reference grid used to reconstruct the high-resolution volume $X$. 
Each stack is generated by a forward model
\begin{equation}
    Y_k = F_k \circ U_k \circ D_k \circ S_k \circ R_k(X) + \eta_k,
    \label{eq:forward}
\end{equation}
where $R_k$ is a rigid transformation accounting for orientation choice and acquisition obliqueness with respect to the canonical reconstruction frame, $S_k$ is a slice-profile convolution along the through-plane axis, $D_k$ applies through-plane downsampling by cropping in the Fourier domain to obtain the anisotropic slice spacing, $U_k$ resamples the resulting stack back to the isotropic grid used for reconstruction and fusion, $F_k$ is a binary \emph{field-of-view} (FOV) operator that masks slices outside the simulated coverage, and $\eta_k$ aggregates k-space and image-domain artefacts (motion ghosting, k-space spikes, aliasing, $B_1^+$ bias fields, Rician noise). 

This formulation is motivated by characteristics of real clinical multi-stack MRI. Example of such clinical aquisition is as shown in \cref{fig:partial_fov} in the appendix.

\subsection{Architecture overview}
\label{sec:method:arch}

MS-HVED adapts the U-HVED hetero-modal architecture~\cite{uhved} to multi-stack super-resolution. As shown in \cref{fig:framework}, the network has three components:
\begin{enumerate}
    \item \textbf{Per-orientation variational encoders.} Each orientation $k\in\{\text{Ax},\text{Cor},\text{Sag}\}$ is processed by an independent encoder $E_{\phi_k}$ with $S=4$ downsampling stages. At each scale $s\in\{0,1,2,3\}$, a $1{\times}1{\times}1$ \emph{variational projection} produces voxel-wise mean $\mu_{k,s}$ and log-variance $\ell_{k,s}=\log\sigma_{k,s}^2$ maps. Each scale defines a local latent Gaussian posterior:
\begin{equation}
    q_{\phi_{k,s}}(z_s \mid Y_k) = \mathcal{N}\!\left(\mu_{k,s},\; \diag(\sigma_{k,s}^2)\right),
    \label{eq:posterior}
\end{equation}
allowing the network to assign spatially-varying confidence based on the anatomical features visible in each view.
    \item \textbf{FOV-aware Product-of-Gaussians fusion.} The per-scale latent posteriors are combined voxel-wise into a fused posterior $q(z_s\mid\mathcal{Y})$ using a precision-weighted product (\Cref{sec:method:fov}).
    \item \textbf{Decoder with multi-scale variational skips.} Samples drawn from the fused latent posterior at every scale are passed to a shared super-resolution decoder $D_\theta$ that progressively upsamples and reconstructs the isotropic HR volume $\hat X$. Optional orientation-specific decoders reconstruct the input stacks $\hat Y_k$.
\end{enumerate}

The encoder and decoder are built from residual blocks without activation normalisation layers, using LeakyReLU activations and residual scaling~\cite{lim2017edsr}. Group, instance, and batch normalisations, which are common in segmentation backbones, are intentionally omitted because they can distort the absolute intensity scale needed for MRI SR. Stability is instead provided by residual scaling, free-bits KL regularisation~\cite{kingma2016iaf}, and spectral normalisation (\Cref{sec:method:spectral}).

\subsection{Spectrally normalised variational head}
\label{sec:method:spectral}

The encoder and decoder are built without activation normalisation layers such as batch, instance or group normalisation. This design choice is motivated by the regression nature of MRI super-resolution, where the network must preserve absolute and relative intensity relationships in the reconstructed image. However, removing activation normalisation also removes a common source of training stabilisation. This is particularly important in a variational architecture, where the encoder predicts voxel-wise Gaussian parameters rather than deterministic features alone.

High-capacity variational encoder-decoders can be difficult to train because the posterior parameterisation must remain both expressive and well regularised. Small perturbations in the predicted log-variance can lead to large changes in the posterior precision and KL divergence~\cite{dai2019}, while overly expressive decoders or poorly balanced KL terms can encourage posterior collapse, where the latent variables carry little information about the input~\cite{wang2023}. We therefore use two complementary stabilisation mechanisms: spectral normalisation~\cite{miyato2018spectral} to constrain the encoders that predict Gaussian parameters, and free-bits KL regularisation~\cite{kingma2016iaf} to reduce posterior collapse.

We apply spectral normalisation to the convolutional layers in the residual pathways and to the $1{\times}1{\times}1$ variational projections. Spectral normalisation acts on the layer weights rather than on the intermediate feature activations. It constrains the operator norm of the corresponding mapping and therefore limits the sensitivity of the predicted posterior mean and log-variance to small perturbations in the input features. Although spectral normalisation was introduced for stabilising GAN discriminators~\cite{miyato2018spectral}, we use it here for its weight-level effect rather than for any adversarial objective.
Importantly, spectral normalisation is not used as a replacement for batch, instance or group normalisation. Those methods normalise feature activations and may alter intensity statistics, whereas spectral normalisation constrains selected weights without normalising the feature maps themselves. 

We also regularise the fused posterior at each latent scale using a free-bits KL objective \cite{kingma2016iaf}. Let
\begin{equation}
\mathrm{KL}_s =
\mathrm{KL}\!\left(q(z_s\mid\mathcal{Y})\,\|\,p(z_s)\right).
\label{eq:kl_scale}
\end{equation}
where $q(z_s\mid\mathcal{Y})$ is the fused posterior at scale $s$ and $p(z_s)=\mathcal{N}(0,I)$ is the prior. The free-bits form is
\begin{equation}
\mathcal{L}_{\textrm{KL},s}
=
\max\!\left(0,\; \mathrm{KL}_s - \beta_{\textrm{fb}}\right),
\label{eq:freebits}
\end{equation}
where $\beta_{\textrm{fb}}$ defines a small tolerance below which the KL term is not penalised. This prevents the objective from forcing informative latent variables to collapse too aggressively towards the prior.

Because the latent feature maps have different spatial sizes across scales, the raw KL magnitude is naturally dominated by the finest scale. We therefore weight each scale-specific KL term by its relative number of latent voxels:
\begin{equation}
\mathcal{L}_{\textrm{KL}}
=
\sum_s
\frac{|z_s|}{\max_{s'} |z_{s'}|}
\mathcal{L}_{\textrm{KL},s}.
\label{eq:kl_multiscale}
\end{equation}
This balances the contribution of the latent pyramid and prevents the finest scale from dominating the variational regularisation.

\subsection{FOV-aware Product-of-Gaussians fusion}
\label{sec:method:fov}

Given conditional independence of stacks under $z_s$ and a weakly informative isotropic prior $p(z_s)=\mathcal{N}(0,I)$, the fused posterior at each scale factorises as a Product-of-Experts, here a Product-of-Gaussians (PoG). The fused precision $T_{\textrm{fused},s}$ and mean $\mu_{\textrm{fused},s}$ are computed voxel-wise:
\begin{align}
    T_{\textrm{fused},s} &= 1 + \sum_{k\in\mathcal{O}} \tilde m_{k,s} \cdot T_{k,s},
    \label{eq:fusprec}\\
    \mu_{\textrm{fused},s} &= \frac{1}{T_{\textrm{fused},s}} \sum_{k\in\mathcal{O}} \tilde m_{k,s} \cdot T_{k,s} \cdot \mu_{k,s},
    \label{eq:fusmean}
\end{align}
where $T_{k,s}=\sigma_{k,s}^{-2}=\exp(-\ell_{k,s})$ and $\tilde m_{k,s}\in[0,1]$ is a \emph{smooth FOV mask} that controls the contribution of orientation $k$ at scale $s$ (see implementation details in \Cref{sec:exp:impl}).

\subsection{Physics-informed domain randomisation}
\label{sec:method:dr}

Synthetic data for supervised training are generated by applying \Cref{eq:forward} to each HR volume in the training set. The simulation includes a trapezoidal slice-profile convolution,  FFT-based cropping for downsampling, motion ghosting, intensity bias field and Rician noise. 

\subsubsection{Obliqueness simulation}
As real clinical multi-stack acquisitions are rarely axis-aligned, we therefore sample a per-stack random Euler rotation
\begin{equation}
    R_k \sim \mathrm{Uniform}\!\left([-\theta_{\max},\theta_{\max}]^3\right), \quad \theta_{\max}=15^\circ,
\end{equation}
and apply $R_k$ to the HR volume's affine matrix before slice-profile convolution and downsampling, so that the resulting LR stack lives on an oblique grid with respect to the canonical reconstruction frame. Crucially, the obliqueness is recorded in the LR stack's affine matrix(and therefore propagates through the resampling step $U_k$). At fusion time, the resampled LR stack appears in the canonical frame as a rotated, partially out-of-bounds volume. This is the same geometry that real oblique acquisitions present after registration to a common reference, and it is the geometry that the FOV mask of \Cref{sec:method:fov} is designed to handle.

We do not explicitly simulate residual registration error in the current implementation. This is a limitation of the present model and is revisited in the discussion.

\subsubsection{Partial-FOV support simulation} 
We simulate partial coverage by suppressing peripheral slices along the through-plane axis while preserving a randomly sampled central coverage fraction. In our experiments the retained fraction is sampled between $40\%$ and $70\%$ of the native LR stack, with a coverage constraint that prevents all stacks from being simultaneously truncated. The resulting binary FOV map $b_k(\mathbf{x})$ is recorded alongside the stack and passed to the FOV-aware fusion module. For evaluation on real clinical inputs, $b_k(\mathbf{x})$ is computed deterministically by resampling the native stack support into the reference grid (see \Cref{sec:method:inference}).

\subsubsection{Other artefacts}
For completeness, the remaining domain-randomisation transforms are: (i)~per-axis spacing sampled uniformly from $[2,9]$\,mm with a thicker through-plane axis, with frequency-domain anti-aliasing baked into the slice-profile filter (ii)~motion ghosting implemented as sinusoidal phase modulation along the k-space phase-encode axis (iii)~k-space spike noise (iv)~mild aliasing (v)~$B_1^+$-style multiplicative bias field (vi)~gamma intensity correction to mimic contrast variability and (vii)~Rician noise.

\subsection{Total training objective}
\label{sec:method:loss}

The full training loss combines the supervised reconstruction terms with the variational and auxiliary orientation-reconstruction terms:
\begin{align}
\mathcal{L}_{\text{total}} 
&= 
w_{\textrm{rec}} \,\mathcal{L}_{\textrm{rec}}
+ w_{\textrm{ssim}} \,\mathcal{L}_{\textrm{SSIM}}
+ w_{\textrm{perc}} \,\mathcal{L}_{\textrm{perc}}
 \nonumber\\
&\quad + 
w_{\textrm{ori}} \,\mathcal{L}_{\textrm{ori}}
+ w_{\textrm{kl}} \sum_s \mathcal{L}_{\textrm{KL},s}
 \nonumber\\
&\quad 
\label{eq:totalloss}
\end{align}
The reconstruction term $\mathcal{L}_{\textrm{rec}}$ is the Charbonnier loss,
\[
\sqrt{(\hat X - X)^2 + \varepsilon^2},
\]
which is a smooth $L_1$ surrogate that improves stability around the optimum compared to plain $L_1$.
The structural term is defined as
$\mathcal{L}_{\textrm{SSIM}} = 1 - \mathrm{SSIM}_{3\mathrm{D}}(\hat X, X)$~\cite{ssim}.
The orientation-specific reconstruction loss term is given by
$\mathcal{L}_{\textrm{ori}} = \sum_k \lVert \hat Y_k - Y_k \rVert_1$.
The KL term corresponds to the free-bits multi-scale KL term defined in \Cref{eq:freebits}.

The perceptual term $\mathcal{L}_{\textrm{perc}}$ is implemented as a 2.5D AlexNet-based feature loss. Perceptual losses compare images in the feature space of a pretrained CNN and better capture visually relevant differences than purely voxel-wise objectives in reconstruction and synthesis~\cite{johnson2016perceptual,zhang2018lpips}, including MRI reconstruction~\cite{ghodrati2019mrreconstruction}. We compute the feature distance on 2D slices sampled from the reconstructed and reference volumes, following recent 2.5D perceptual supervision for volumetric medical images~\cite{kim2024cyclic25d}, because fully 3D perceptual losses are memory intensive and require suitable pretrained 3D feature extractors. AlexNet provides a lightweight perceptual backbone commonly used in LPIPS-style metrics~\cite{zhang2018lpips}. This term is used as an auxiliary loss alongside Charbonnier, SSIM and KL terms to encourage sharper structural detail without replacing voxel-wise fidelity constraints. 

\subsection{Inference}
\label{sec:method:inference}

At inference time, given a set of clinical stacks $\{Y_k\}_{k\in\mathcal{O}}$, MS-HVED first computes a per-orientation FOV map $b_k(\mathbf{x})$ by registering and resampling the native stack into a reference grid. This captures both limited stack coverage and obliqueness-induced out-of-bounds regions after registration. Stacks and FOV maps are padded to match the encoder's downsampling factor. The encoders are then applied independently to each orientation, the FOV mask of \Cref{eq:smoothfov} is computed per scale, and \Cref{eq:fusprec,eq:fusmean} produce the fused posterior. We use the posterior mean for deterministic reconstruction.

\section{Experiments}
\label{sec:experiments}

\subsection{Evaluation design}
\label{sec:exp:design}

The evaluation was designed to test five complementary properties of MS-HVED:
\begin{enumerate}
    \item Reconstruction fidelity relative to single-stack, self-supervised, and fixed-input supervised baselines.
    \item The ability of one trained model to accept arbitrary non-empty subsets of axial, coronal, and sagittal stacks.
    \item The effect of dataset-specific versus sequence-agnostic training.
    \item The contributions of oblique partial-FOV simulation and FOV-aware fusion.
    \item Generalisation to real multi-stack acquisitions.
\end{enumerate}
The first four questions were evaluated using synthetic data with known HR targets. Real-data performance was assessed using both a paired LR--HR cohort and a blinded expert-reader study using seven board-certified neuroradiologists.

\subsection{Datasets}
\label{sec:exp:datasets}

\subsubsection{Source MRI cohorts}

Three high-resolution (HR) brain MRI cohorts were used as source data for training and synthetic evaluation. The first was an internal meningioma cohort (\KCHDAT) comprising $774$ T1-weighted volumes acquired under heterogeneous clinical protocols. The second and third were the publicly available IXI T1-weighted and T2-weighted cohorts~\cite{ixi}. The internal cohort was split into $537/110/127$ training/validation/test cases, while the IXI T1-weighted and T2-weighted cohorts comprised $406/88/87$ and $404/88/86$ training/validation/test volumes, respectively.

The internal clinical data were accessed under the AI-driven Management of Brain Tumours (AIMBraTS) study, approved by the NHS Research Ethics Committee (REC reference: 22/NS/0160; IRAS ID: 320754). Individual patient consent was not required because the study involved retrospective analysis of routinely collected clinical data. All internal data were handled in accordance with the UK Data Protection Act 2018~\cite{DPA2018} and UK General Data Protection Regulation~\cite{UKGDPR2018}. The IXI dataset is publicly available and was used in accordance with its data-access conditions.

All HR volumes were defaced~\cite{defacing}, resampled to $1 \mathrm{mm}$ isotropic resolution using trilinear interpolation, and intensity-normalised to $[0,1]$.

\subsubsection{Synthetic LR--HR data}

The HR volumes described above were used to generate synthetic anisotropic LR stacks using the physics-informed forward model in \Cref{sec:method}. For each HR volume, axial, coronal and sagittal stacks were generated with randomised acquisition parameters, including anisotropic resolution, oblique prescription, partial-FOV coverage and image artefacts. The original HR volume served as the reference target for supervised training or full-reference quantitative evaluation depending on which split they were assigned to.

\subsubsection{Real evaluation cohorts}
To assess generalisation to real multi-stack acquisitions, we evaluated MS-HVED on three additional cohorts. The first consisted of $7$ paired LR--HR T2-weighted healthy-control brain MRI cases from \citet{Ebner2019}. The HR acquisition is used as a pseudo-reference for quantitative evaluation, because the LR and HR scans are necessarily acquired with different acquisition settings. Consequently, differences in contrast, intensity scale and image appearance may reflect acquisition-physics mismatch rather than reconstruction error alone. This $7$ cases were also included in the expert-reader study, allowing the reference-based and subjective assessments to be interpreted together.

The second cohort consisted of $7$ low-resolution multi-stack meningioma cases from our institution, acquired under the AIMBraTS study approval described above (REC reference: 22/NS/0160; IRAS ID: 320754). These cases contain clinically relevant tumour pathology but do not have matched 3D HR reference acquisitions. In this cohort, $6$ of the $7$ cases contain only two orientations, providing a realistic test for missing orientation.

The third cohort was drawn from the publicly available Nigerian clinical MRI dataset~\cite{wogu2025nigerian,nigerianbrainlife2024}. The full dataset contains scans from $35$ healthy participants, $31$ individuals with age-related dementia and $22$ individuals with Parkinson's disease, with T1-weighted, T2-weighted, FLAIR and DWI acquisitions. We used $30$ cases (consisting of T1-weighted, T2-weighted and FLAIR images) from this cohort as an additional reference-free clinical generalisation test, reflecting acquisition and population characteristics distinct from the training cohorts.

For all real evaluation cohorts, the available LR stacks were rigidly registered to a common reference space and resampled to $1,\mathrm{mm}$ isotropic resolution. When no matched HR reference was available, the reconstruction grid was defined by the union of the available LR stack supports.

\subsection{Implementation details}
\label{sec:exp:impl}

MS-HVED is implemented in PyTorch~\cite{pytorch} and trained on a single NVIDIA A100 GPU. We use the AdamW~\cite{adamW} optimiser with learning rate $10^{-4}$ and weight decay $10^{-5}$, a cosine annealing schedule with $5\%$ linear warm-up, and mixed-precision training. Training is run for up to $1000$ epochs with early stopping, using patches of size $128{\times}128{\times}64$ and a batch size of $2$. Free-bits is set to $\beta_{\textrm{fb}}{=}1$ nat per scale-aggregated KL term. Spectral normalisation is applied to the convolutional residual pathways and to the variational projections.
Default loss weights are fixed across all experiments:
\begin{equation*}
\begin{aligned}
w_{\textrm{rec}} &= 0.4, \quad
w_{\textrm{ori}} = 0.4, \quad
w_{\textrm{kl}} = 0.1, \\
w_{\textrm{ssim}} &= 0.1, \quad
w_{\textrm{perc}} = 0.1.
\end{aligned}
\end{equation*}
Domain-randomisation transforms are implemented using PyTorch~\cite{pytorch} and MONAI~\cite{monai}.

To evaluate sequence-agnostic training, we trained three dataset-specific MS-HVED models independently on the IXI-T1, IXI-T2 and \KCHDAT{} training sets, and a fourth model on the combined training data. The sequence-agnostic model therefore encountered both T1- and T2-weighted images and data from different acquisition cohorts during training, while retaining the same architecture, loss function and optimisation settings as the dataset-specific models. At evaluation, the shared model was tested separately on each held-out cohort and compared with the corresponding dataset-specific model.

For FOV-aware fusion, each stack is associated with a binary mask $b_k(\mathbf{x})\in{0,1}$, where $1$ denotes voxels inside the spatial coverage of orientation $k$ and $0$ denotes voxels outside it. For synthetic data, this support map is obtained directly from the simulated stack geometry. For real data, it is computed by resampling the native stack support into the common reconstruction grid. At each latent scale $s$, the binary mask is downsampled and smoothed with an isotropic Gaussian kernel:
\begin{equation}
\tilde m_{k,s}(\mathbf{x})
=
\left(g_\sigma * b_k\right)(\mathbf{x}_s),
\label{eq:smoothfov}
\end{equation}
where $g_\sigma$ has standard deviation $\sigma{=}3$ voxels and $\mathbf{x}_s$ denotes the coordinate at scale $s$. Global orientation availability is handled by multiplying $\tilde m_{k,s}$ by an orientation-level indicator, so absent stacks contribute zero precision during POG fusion.

\subsection{Baselines and quantitative metrics}
\label{sec:exp:baselines}

\subsubsection{Baselines}
Synthetic comparisons included ECLARE~\cite{eclare}, TripleSR~\cite{triplesr}, and a supervised SegResNet~\cite{segresnet}. SegResNet was trained separately for each dataset and input configuration (\textsc{Ax}, \textsc{Ax+Cor}, and \textsc{Ax+Cor+Sag}), whereas a single MS-HVED model per training setting accepted every non-empty orientation subset. On the paired real cohort, we additionally evaluated NiftyMIC~\cite{Ebner2019,NiftyMiC} as a representative classical approach and a naive baseline in the form of a non-learned FOV-weighted average (Naive+FOV) computed by:
\begin{equation}
\hat X_{\mathrm{mean+FOV}}(\mathbf{x})=
\frac{\sum_{k\in\mathcal O}\tilde m_k(\mathbf{x})Y_k(\mathbf{x})}
{\sum_{k\in\mathcal O}\tilde m_k(\mathbf{x})},
\label{eq:naive_fov_recon}
\end{equation}
Here, $Y_k$ is the registered stack and $\tilde m_k$ is its FOV-derived contribution weight.

\subsubsection{Quantitative metrics}
Quantitative evaluation of MRI super-resolution is not straightforward because different metrics capture different aspects of reconstruction quality. Voxel-wise errors such as mean absolute error (MAE) and peak signal-to-noise ratio (PSNR) measure numerical agreement with a reference image, but they may favour smooth reconstructions and can be sensitive to small contrast differences. 3D structural similarity (SSIM)~\cite{ssim} provides a complementary measure of local structural similarity, while perceptual losses are more sensitive to image sharpness and feature-level differences. For this reason, we report multiple metrics rather than relying on a single score.

On synthetic data, we report MAE, PSNR, SSIM, and the 2.5D AlexNet-based perceptual loss. Metrics are computed over the HR target region captured by the input stacks used by each method, as described below:

Let $M_o$ denote the FOV mask for orientation $o\in\{\textsc{Ax},\textsc{Cor},\textsc{Sag}\}$. For a method using the orientation set $\mathcal{O}$, metrics are computed over
\[
    M_{\mathcal{O}} = \bigcup_{o\in\mathcal{O}} M_o .
\]
Thus, single-stack axial methods are evaluated only over $M_{\textsc{Ax}}$, dual-stack methods such as TripleSR and MS-HVED \textsc{Ax+Cor} are evaluated over $M_{\textsc{Ax}}\cup M_{\textsc{Cor}}$, and three-stack methods are evaluated over $M_{\textsc{Ax}}\cup M_{\textsc{Cor}}\cup M_{\textsc{Sag}}$, corresponding to the full HR target region captured by the available synthetic acquisitions. This avoids penalising a method for anatomy outside its input FOV while still evaluating reconstruction fidelity over the region it could in principle reconstruct.

On the paired healthy-control real cohort, we report PSNR, SSIM, normalised cross-correlation (NCC), and perceptual loss against the HR pseudo-reference. Since the LR and HR images were acquired separately and may differ in intensity scale, contrast and scanner-dependent appearance, metrics are computed after linear calibration of each reconstruction to the HR pseudo-reference using ordinary least squares,
\[
    \hat x_{\textrm{cal}} = a\hat x + b,
\]
where $a$ and $b$ are estimated by ordinary least squares. This calibration removes global linear intensity differences, allowing PSNR and SSIM to better reflect spatial and structural agreement. NCC is also reported because it is insensitive to global linear intensity variations.

To assess whether image-level improvements translate to improved downstream analysis, we evaluate tissue segmentation performance on the synthetic test sets. SynthSeg~\cite{SynthSeg} is applied to the axial LR inputs, SegResNet reconstructions and MS-HVED reconstructions. Segmentations obtained from the corresponding HR volumes are used as reference segmentations. We evaluate cerebrospinal fluid (CSF), grey matter (GM), and white matter (WM) using Dice score and the 95th-percentile Hausdorff distance (HD$_{95}$, in mm). This provides a task-oriented evaluation of whether the reconstructed images preserve anatomical structures relevant to automated brain analysis.

\subsection{Expert-reader study}
\label{sec:exp:expert_review}
Reference-free clinical evaluation is necessary for the real cohorts because no isotropic HR reference is available. In these cohorts, image quality is assessed by expert review using a blinded side-by-side rating protocol. Seven board-certified neuroradiologists were shown SegResNet and MS-HVED reconstructions in random order with method labels hidden. Cases were reviewed on a dedicated interface that allows multi-planar inspection of the reconstructed volumes and comparison against the available LR inputs. A screenshot of this interface is provided in \Cref{fig:reviewer_ui}.

Experts score each reconstruction on a $1$--$9$ ordinal scale for the following criteria: freedom from artefact, signal-to-noise ratio (SNR), sharpness, visibility of tumour pathology where applicable, and overall image quality. Reconstructions were assessed in pairs, and readers were not permitted to assign the same score to both methods for a given case and criterion, thereby requiring a directional preference while retaining the ordinal rating scale. This introduces a forced choice component analogous to pairwise and two-alternative forced choice (2AFC) assessment~\cite{gur1997forcedchoice}. All criteria were scored with higher values indicating better image quality. The pathology-visibility criterion was applied only to cases with visible focal pathology, i.e. the meningioma cohort, and was marked not applicable for the others.

Input availability was tracked separately from cohort. In some cases, one or more stacks were unavailable, so they could not be used for reconstruction or displayed in the review interface. In other cases, an axial or coronal stack was available and remained visible to the reader in the interface, but was deliberately omitted from both reconstruction methods to test performance with a missing input. For each case, we recorded which orientation stacks were available, which were used for reconstruction, and whether an unused stack was naturally unavailable or deliberately omitted. Results were then grouped according to the inputs used for reconstruction: full input, axial omitted, or coronal omitted.

Scores are reported descriptively as mean$\pm$standard deviation. Within each cohort or input group, MS-HVED and SegResNet ratings from the same reader and case were compared using two-sided Wilcoxon signed-rank tests at $\alpha=0.05$.  Reader-specific use of the $1$--$9$ scale was addressed with a scale-invariant paired-preference summary. For reader $r$, cohort $c$, and criterion $q$, we defined
\begin{equation}
P_{r,c,q}=100\,\frac{W_{r,c,q}-L_{r,c,q}}
{W_{r,c,q}+L_{r,c,q}},
\label{eq:reader_net_preference}
\end{equation}
where $W$ and $L$ are the numbers of case-level pairs for which MS-HVED received a higher or lower score than SegResNet, respectively. $P=0$ denotes balanced preference, positive values favour MS-HVED, and negative values favour SegResNet. 

Inter-rater variability is summarised for shared case--method--criterion units using pairwise quadratic-weighted $\kappa$, the proportion of ratings within one point, and the mean absolute score difference. 

\section{Results}
\label{sec:results}

\subsection{Synthetic dataset evaluation}
\label{sec:res:oblique_synth}

\begin{figure*}[!t]
    \centering
    \includegraphics[width=\textwidth]{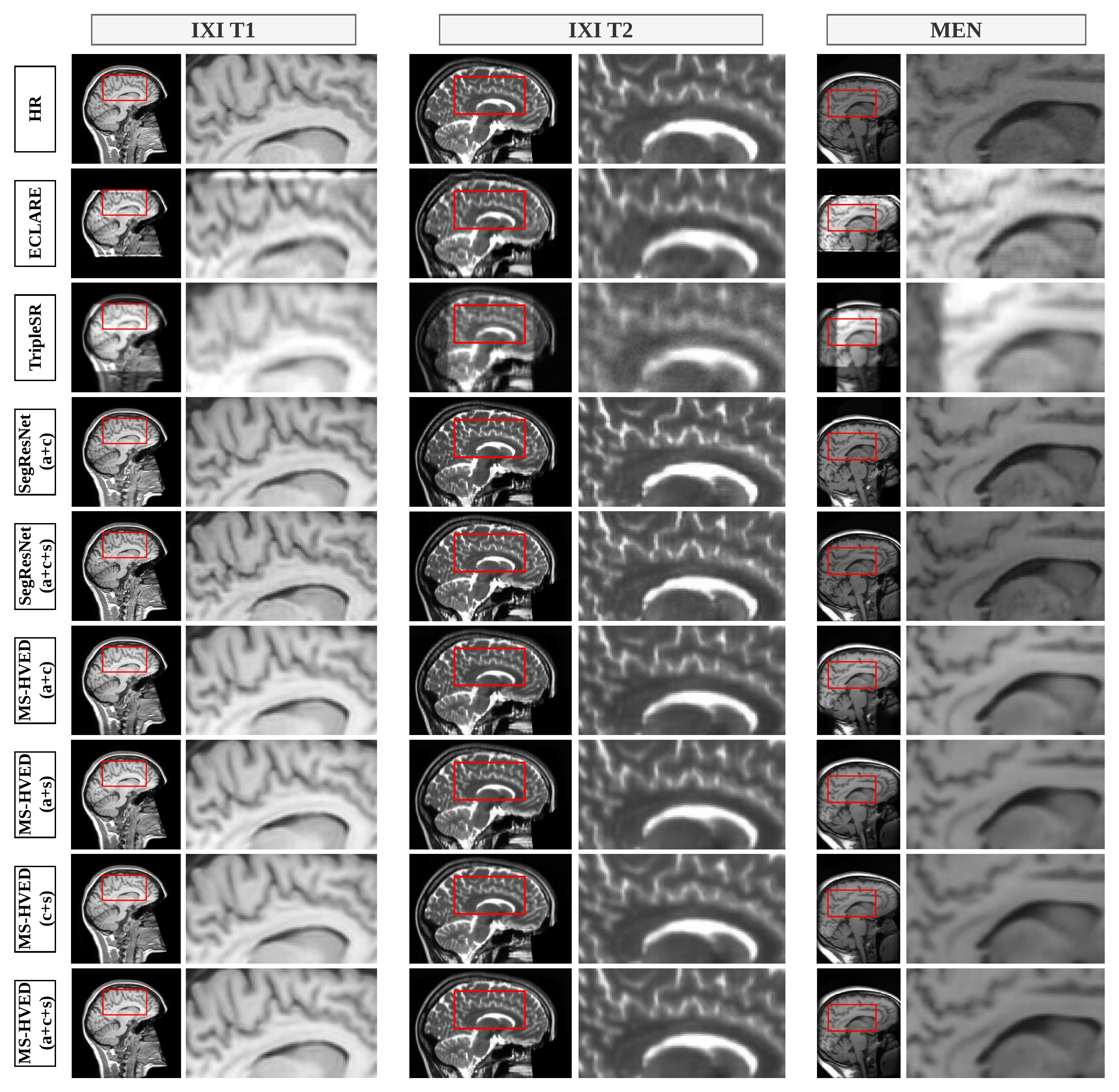}
    \caption{Reference HR image and qualitative SR results from the IXI\cite{ixi} and \KCHDAT synthetic benchmark datasets. The presented results were reconstructed from $5mm$ anisotropic inputs.}
    \label{fig:qualitative}
\end{figure*}

As summarized in \Cref{tab:results_oblique_synth}, MS-HVED was competitive across all three synthetic test cohorts. With three inputs, it obtained the best MAE, PSNR, SSIM, and perceptual loss on \KCHDAT. It also achieves the best MAE, PSNR, and SSIM on IXI-T1, with perceptual loss tied, and the second-best values on IXI-T2, where three-input SegResNet performed best. When tested on two inputs, MS-HVED still remained competitive with fixed-input baselines, particularly for SSIM on \KCHDAT and IXI-T1. Representative synthetic reconstructions are shown in \Cref{fig:qualitative}.

\begin{table*}[!t]
\centering
\caption{\textbf{Synthetic benchmark.} All methods are evaluated on the same oblique test inputs, with metrics computed only over the HR target region supported by the stack orientations used by each method. Values report mean{\scriptsize$\pm$}standard deviation. Best values within each dataset and metric are bolded, and second-best values are underlined.}
\label{tab:results_oblique_synth}
\small
\setlength{\tabcolsep}{3.5pt}
\begin{tabular}{llcccc}
\toprule
\textbf{Data} & \textbf{Method} & \textbf{MAE}$\downarrow$ & \textbf{PSNR}$\uparrow$ & \textbf{SSIM}$\uparrow$ & \textbf{Percept.}$\downarrow$ \\
\midrule
\multirow{7}{*}{\KCHDAT}
& ECLARE~\cite{eclare}             & .083{\scriptsize$\pm$.033} & 16.39{\scriptsize$\pm$3.8} & .57{\scriptsize$\pm$.16} & .64{\scriptsize$\pm$.25} \\
& TripleSR~\cite{triplesr}         & .057{\scriptsize$\pm$.017} & 19.43{\scriptsize$\pm$2.4} & .68{\scriptsize$\pm$.12} & .55{\scriptsize$\pm$.18} \\
& SegResNet (Ax)                   & .035{\scriptsize$\pm$.011} & 24.02{\scriptsize$\pm$2.5} & .81{\scriptsize$\pm$.08} & .17{\scriptsize$\pm$.06} \\
& SegResNet (Ax+Cor)               & .032{\scriptsize$\pm$.008} & 24.79{\scriptsize$\pm$2.3} & .81{\scriptsize$\pm$.08} & \underline{.14{\scriptsize$\pm$.05}} \\
& SegResNet (Ax+Cor+Sag)           & .034{\scriptsize$\pm$.011} & 24.44{\scriptsize$\pm$2.5} & \underline{.82{\scriptsize$\pm$.10}} & \underline{.14{\scriptsize$\pm$.06}} \\
& MS-HVED (Ax+Cor)               & \underline{.028{\scriptsize$\pm$.008}} & \underline{25.43{\scriptsize$\pm$2.3}} & \textbf{.88{\scriptsize$\pm$.04}} & .16{\scriptsize$\pm$.07} \\
& MS-HVED (Ax+Cor+Sag)           & \textbf{.023{\scriptsize$\pm$.006}} & \textbf{27.01{\scriptsize$\pm$2.1}} & \textbf{.88{\scriptsize$\pm$.04}} & \textbf{.13{\scriptsize$\pm$.05}} \\
\midrule
\multirow{7}{*}{IXI-T1}
& ECLARE~\cite{eclare}             & .093{\scriptsize$\pm$.036} & 15.93{\scriptsize$\pm$3.4} & .57{\scriptsize$\pm$.15} & .71{\scriptsize$\pm$.24} \\
& TripleSR~\cite{triplesr}         & .058{\scriptsize$\pm$.021} & 20.02{\scriptsize$\pm$3.3} & .64{\scriptsize$\pm$.14} & .55{\scriptsize$\pm$.22} \\
& SegResNet (Ax)                   & .032{\scriptsize$\pm$.008} & 24.39{\scriptsize$\pm$2.1} & .88{\scriptsize$\pm$.05} & \underline{.14{\scriptsize$\pm$.04}} \\
& SegResNet (Ax+Cor)               & .029{\scriptsize$\pm$.008} & 25.12{\scriptsize$\pm$2.1} & \underline{.91{\scriptsize$\pm$.04}} & \underline{.14{\scriptsize$\pm$.05}} \\
& SegResNet (Ax+Cor+Sag)           & \underline{.027{\scriptsize$\pm$.007}} & \underline{26.24{\scriptsize$\pm$2.1}} & .86{\scriptsize$\pm$.09} & \textbf{.12{\scriptsize$\pm$.05}} \\
& MS-HVED (Ax+Cor)               & \underline{.027{\scriptsize$\pm$.009}} & 25.75{\scriptsize$\pm$2.5} & \underline{.91{\scriptsize$\pm$.04}} & .15{\scriptsize$\pm$.07} \\
& MS-HVED (Ax+Cor+Sag)           & \textbf{.023{\scriptsize$\pm$.007}} & \textbf{27.16{\scriptsize$\pm$2.3}} & \textbf{.93{\scriptsize$\pm$.03}} & \textbf{.12{\scriptsize$\pm$.05}} \\
\midrule
\multirow{7}{*}{IXI-T2}
& ECLARE~\cite{eclare}             & .066{\scriptsize$\pm$.024} & 17.80{\scriptsize$\pm$3.6} & .64{\scriptsize$\pm$.14} & .47{\scriptsize$\pm$.19} \\
& TripleSR~\cite{triplesr}         & .045{\scriptsize$\pm$.010} & 21.17{\scriptsize$\pm$1.9} & .65{\scriptsize$\pm$.18} & .46{\scriptsize$\pm$.13} \\
& SegResNet (Ax)                   & .029{\scriptsize$\pm$.007} & 24.89{\scriptsize$\pm$2.1} & .87{\scriptsize$\pm$.04} & .12{\scriptsize$\pm$.05} \\
& SegResNet (Ax+Cor)               & .023{\scriptsize$\pm$.005} & 26.26{\scriptsize$\pm$1.7} & \textbf{.91{\scriptsize$\pm$.03}} & .11{\scriptsize$\pm$.04} \\
& SegResNet (Ax+Cor+Sag)           & \textbf{.019{\scriptsize$\pm$.004}} & \textbf{28.42{\scriptsize$\pm$1.7}} & \underline{.88{\scriptsize$\pm$.07}} & \textbf{.09{\scriptsize$\pm$.03}} \\
& MS-HVED (Ax+Cor)               & .024{\scriptsize$\pm$.005} & 26.47{\scriptsize$\pm$1.7} & .87{\scriptsize$\pm$.06} & .12{\scriptsize$\pm$.04} \\
& MS-HVED (Ax+Cor+Sag)           & \underline{.021{\scriptsize$\pm$.004}} & \underline{27.71{\scriptsize$\pm$1.5}} & \underline{.88{\scriptsize$\pm$.06}} & \underline{.10{\scriptsize$\pm$.03}} \\
\bottomrule
\end{tabular}
\end{table*}

\subsubsection{Flexible orientation handling}
\label{sec:res:orientation}

The same MS-HVED model was evaluated with two additional dual-input combinations. As shown in \Cref{tab:flexi_orient}, performance was similar for \textsc{Ax+Sag} and \textsc{Cor+Sag} on every test cohort. Across the three cohorts, the absolute PSNR difference between these input combinations was at most $0.75$~dB. These results support the intended use of one model with different available orientation subsets, without retraining a separate network for each subset.

\begin{table}[!t]
\centering
\caption{\textbf{MS-HVED performance for additional dual-orientation input combinations.} Both configurations were evaluated using the same trained model for each synthetic benchmark dataset. Values are mean{\scriptsize$\pm$}standard deviation.}
\label{tab:flexi_orient}
\small
\setlength{\tabcolsep}{3pt}
\begin{tabular}{llcccc}
\toprule
\textbf{Data} & \textbf{Orientations} & \textbf{MAE}$\downarrow$ & \textbf{PSNR}$\uparrow$ & \textbf{SSIM}$\uparrow$ & \textbf{Percept.}$\downarrow$ \\
\midrule
\multirow{2}{*}{\KCHDAT} & Ax+Sag  & .027{\scriptsize$\pm$.008} & 25.69{\scriptsize$\pm$2.3} & .87{\scriptsize$\pm$.04} & .16{\scriptsize$\pm$.06} \\
                        & Cor+Sag & .026{\scriptsize$\pm$.007} & 26.17{\scriptsize$\pm$2.1} & .86{\scriptsize$\pm$.04} & .16{\scriptsize$\pm$.06} \\
\midrule
\multirow{2}{*}{IXI-T1} & Ax+Sag  & .027{\scriptsize$\pm$.008} & 25.67{\scriptsize$\pm$2.4} & .91{\scriptsize$\pm$.04} & .15{\scriptsize$\pm$.06} \\
                       & Cor+Sag & .026{\scriptsize$\pm$.008} & 26.42{\scriptsize$\pm$2.4} & .91{\scriptsize$\pm$.03} & .14{\scriptsize$\pm$.06} \\
\midrule
\multirow{2}{*}{IXI-T2} & Ax+Sag  & .024{\scriptsize$\pm$.005} & 26.50{\scriptsize$\pm$1.8} & .89{\scriptsize$\pm$.05} & .12{\scriptsize$\pm$.04} \\
                       & Cor+Sag & .024{\scriptsize$\pm$.005} & 26.63{\scriptsize$\pm$1.7} & .88{\scriptsize$\pm$.03} & .13{\scriptsize$\pm$.04} \\
\bottomrule
\end{tabular}
\end{table}

\subsubsection{Sequence-agnostic reconstruction}
\label{sec:res:sequence_agnostic}

As shown in \Cref{tab:sequence_agnostic}, sequence-agnostic training remained competitive with dataset-specific training. It improved PSNR and perceptual loss on IXI-T1 and \KCHDAT, and achieved the lowest perceptual loss on IXI-T2. The dataset-specific models retained higher SSIM on IXI-T1 and IXI-T2 and higher PSNR on IXI-T2. Thus, a shared model reduced the need for contrast/sequence specific training while maintaining comparable performance across cohorts.

\begin{table}[!t]
\centering
\caption{\textbf{Data-specific (DS) vs sequence-agnostic (SA) MS-HVED} Each row reports performance on the held-out test set of the named cohort in the synthetic benchmark. Best values for each cohort and metric are bolded.}
\label{tab:sequence_agnostic}
\small
\setlength{\tabcolsep}{2pt}

\begin{tabular}{llcccc}
\toprule
\textbf{Data} & \textbf{Model} & \textbf{MAE}$\downarrow$ & \textbf{PSNR}$\uparrow$ & \textbf{SSIM}$\uparrow$ & \textbf{Percept.}$\downarrow$ \\
\midrule

\multirow{2}{*}{IXI-T1} 
& DS       
& \textbf{.023{\scriptsize$\pm$.007}} 
& 27.16{\scriptsize$\pm$2.3} 
& \textbf{.928{\scriptsize$\pm$.027}} 
& .121{\scriptsize$\pm$.049} \\

& SA 
& .024{\scriptsize$\pm$.007} 
& \textbf{27.36{\scriptsize$\pm$2.4}} 
& .878{\scriptsize$\pm$.075} 
& \textbf{.110{\scriptsize$\pm$.048}} \\

\midrule
\multirow{2}{*}{IXI-T2} 
& DS      
& \textbf{.021{\scriptsize$\pm$.004}} 
& \textbf{27.71{\scriptsize$\pm$1.5}} 
& \textbf{.883{\scriptsize$\pm$.057}} 
& .102{\scriptsize$\pm$.034} \\

& SA 
& .024{\scriptsize$\pm$.004} 
& 27.09{\scriptsize$\pm$1.3} 
& .825{\scriptsize$\pm$.110} 
& \textbf{.099{\scriptsize$\pm$.033}} \\

\midrule
\multirow{2}{*}{\KCHDAT} 
& DS   
& .023{\scriptsize$\pm$.006} 
& 27.01{\scriptsize$\pm$2.1} 
& \textbf{.883{\scriptsize$\pm$.037}} 
& .133{\scriptsize$\pm$.053} \\

& SA 
& \textbf{.022{\scriptsize$\pm$.006}} 
& \textbf{27.45{\scriptsize$\pm$2.2}} 
& .882{\scriptsize$\pm$.058} 
& \textbf{.118{\scriptsize$\pm$.046}} \\

\bottomrule
\end{tabular}
\end{table}

\subsubsection{Ablation study}
\label{sec:res:fov_oblique_ablation}

We ablate the two geometric components introduced to make MS-HVED robust to realistic multi-stack acquisition settings: oblique and partial-FOV of simulated datasets, and FOV-aware mask in PoG fusion. The first component exposes the model to stacks whose spatial supports are rotated, incomplete and only partially overlapping, matching the geometry commonly observed in clinical multi-orientation acquisitions. The second component provides voxel-wise information during fusion, allowing the model to distinguish between valid low-resolution evidence and regions where an orientation does not contribute anatomical information.

As presented in \cref{tab:fov_oblique_ablation}, oblique partial-FOV simulation produced the largest improvement, and FOV-mask conditioning provided a further consistent benefit. Relative to the variant with neither component, the complete model improved PSNR by $6.06$, $5.89$, and $4.46$~dB on \KCHDAT, IXI-T1, and IXI-T2, respectively. Adding the masks to the FOV-simulated model further improved SSIM by $.014$--$.049$ and reduced perceptual loss by $.031$--$.039$.

\begin{table*}[pos=!tbp]
\centering
\caption{\textbf{Ablation of oblique partial-FOV simulation and FOV-aware mask conditioning.}  All variants are evaluated on the same held-out synthetic test sets with oblique partial-FOV geometry. Check marks indicate whether each component is used. Values report mean{\scriptsize$\pm$}standard deviation.}
\label{tab:fov_oblique_ablation}
\small
\setlength{\tabcolsep}{2pt}
\renewcommand{\arraystretch}{1.08}
\begin{tabular}{lcc cccc}
\toprule
\textbf{Data} 
& \textbf{FOV sim.} 
& \textbf{FOV masks} 
& \textbf{MAE}$\downarrow$ 
& \textbf{PSNR}$\uparrow$ 
& \textbf{SSIM}$\uparrow$ 
& \textbf{Percept.}$\downarrow$ \\
\midrule

\multirow{3}{*}{\KCHDAT}
& \xmark & \xmark
& .043{\scriptsize$\pm$.014}
& 20.95{\scriptsize$\pm$2.76}
& .821{\scriptsize$\pm$.057}
& .207{\scriptsize$\pm$.075} \\

& \cmark & \xmark
& .024{\scriptsize$\pm$.007}
& 26.18{\scriptsize$\pm$2.06}
& .869{\scriptsize$\pm$.044}
& .172{\scriptsize$\pm$.071} \\

& \cmark & \cmark
& \textbf{.023{\scriptsize$\pm$.006}}
& \textbf{27.01{\scriptsize$\pm$2.1}}
& \textbf{.883{\scriptsize$\pm$.037}}
& \textbf{.133{\scriptsize$\pm$.053}} \\

\midrule

\multirow{3}{*}{IXI-T1}
& \xmark & \xmark
& .045{\scriptsize$\pm$.015}
& 21.27{\scriptsize$\pm$2.91}
& .810{\scriptsize$\pm$.087}
& .195{\scriptsize$\pm$.089} \\

& \cmark & \xmark
& .025{\scriptsize$\pm$.008}
& 26.79{\scriptsize$\pm$2.57}
& .879{\scriptsize$\pm$.054}
& .157{\scriptsize$\pm$.082} \\

& \cmark & \cmark
& \textbf{.023{\scriptsize$\pm$.007}}
& \textbf{27.16{\scriptsize$\pm$2.3}}
& \textbf{.928{\scriptsize$\pm$.027}}
& \textbf{.121{\scriptsize$\pm$.049}} \\

\midrule

\multirow{3}{*}{IXI-T2}
& \xmark & \xmark
& .034{\scriptsize$\pm$.008}
& 23.25{\scriptsize$\pm$2.07}
& .784{\scriptsize$\pm$.091}
& .167{\scriptsize$\pm$.059} \\

& \cmark & \xmark
& \textbf{.021{\scriptsize$\pm$.004}}
& 27.52{\scriptsize$\pm$1.54}
& .842{\scriptsize$\pm$.074}
& .133{\scriptsize$\pm$.052} \\

& \cmark & \cmark
& \textbf{.021{\scriptsize$\pm$.004}}
& \textbf{27.71{\scriptsize$\pm$1.5}}
& \textbf{.883{\scriptsize$\pm$.057}}
& \textbf{.102{\scriptsize$\pm$.034}} \\

\bottomrule
\end{tabular}
\end{table*}

\subsubsection{Downstream evaluation}
\label{sec:res:downstream}

We assess downstream utility using SynthSeg~\cite{SynthSeg} on the synthetic test cases, treating the segmentations of the HR volumes as reference segmentations. Dice score and the 95th-percentile Hausdorff distance (HD$_{95}$, in mm) for CSF, GM, and WM are presented in \cref{tab:downstream_seg}, with a representative case visualized in the appendix (\cref{fig:downstream_seg}).

Both learned reconstructions markedly improved SynthSeg performance relative to a single LR input (\Cref{tab:downstream_seg}). MS-HVED and SegResNet had nearly identical Dice scores on IXI-T1 and IXI-T2. On \KCHDAT, MS-HVED improved CSF Dice from .86 to .87 and reduced HD$_{95}$ for CSF, GM, and WM. 

\begin{table*}[pos=!tbp]
\centering
\caption{\textbf{Downstream brain tissue segmentation on the synthetic test sets.} HR-volume segmentations served as reference. Best values within each cohort and metric are bolded, and second-best values are underlined. Ties at the displayed precision are marked equally.}
\label{tab:downstream_seg}
\footnotesize
\setlength{\tabcolsep}{3.5pt}

\begin{tabular}{l l cccccc}
\toprule
\multirow{2}{*}{} & \multirow{2}{*}{Method} & \multicolumn{2}{c}{CSF} & \multicolumn{2}{c}{GM} & \multicolumn{2}{c}{WM} \\
\cmidrule(lr){3-4} \cmidrule(lr){5-6} \cmidrule(lr){7-8}
& & Dice$\uparrow$ & HD$_{95}\downarrow$ & Dice$\uparrow$ & HD$_{95}\downarrow$ & Dice$\uparrow$ & HD$_{95}\downarrow$ \\
\midrule
\multirow{3}{*}{\KCHDAT}
& LR Input 
& $0.58{\pm}0.20$ 
& $12.10{\pm}28.41$ 
& $0.69{\pm}0.18$ 
& $9.91{\pm}25.56$ 
& $0.75{\pm}0.18$ 
& $9.67{\pm}25.17$ \\
& SegResNet 
& $\underline{0.86{\pm}0.05}$ 
& $\underline{1.21{\pm}0.36}$ 
& $\mathbf{0.90{\pm}0.04}$ 
& $\underline{1.13{\pm}0.35}$ 
& $\mathbf{0.93{\pm}0.03}$ 
& $\underline{1.18{\pm}0.41}$ \\
& MS-HVED 
& $\mathbf{0.87{\pm}0.04}$ 
& $\mathbf{1.18{\pm}0.26}$ 
& $\mathbf{0.90{\pm}0.03}$ 
& $\mathbf{1.06{\pm}0.17}$ 
& $\mathbf{0.93{\pm}0.02}$ 
& $\mathbf{1.08{\pm}0.21}$ \\
\midrule
\multirow{3}{*}{IXI-T1}
& LR Input 
& $0.60{\pm}0.18$ 
& $7.18{\pm}8.67$ 
& $0.73{\pm}0.16$ 
& $5.47{\pm}8.38$ 
& $0.78{\pm}0.16$ 
& $5.62{\pm}8.42$ \\
& SegResNet 
& $\mathbf{0.87{\pm}0.05}$ 
& $\mathbf{1.14{\pm}0.20}$ 
& $\mathbf{0.91{\pm}0.03}$ 
& $\mathbf{1.05{\pm}0.14}$ 
& $\mathbf{0.94{\pm}0.02}$ 
& $\underline{1.07{\pm}0.16}$ \\
& MS-HVED 
& $\underline{0.86{\pm}0.05}$ 
& $\underline{1.16{\pm}0.20}$ 
& $\mathbf{0.91{\pm}0.03}$ 
& $\mathbf{1.05{\pm}0.14}$ 
& $\mathbf{0.94{\pm}0.02}$ 
& $\mathbf{1.06{\pm}0.15}$ \\
\midrule
\multirow{3}{*}{IXI-T2}
& LR Input 
& $0.57{\pm}0.17$ 
& $8.22{\pm}6.90$ 
& $0.70{\pm}0.15$ 
& $5.46{\pm}5.41$ 
& $0.78{\pm}0.12$ 
& $5.61{\pm}5.40$ \\
& SegResNet 
& $\mathbf{0.86{\pm}0.04}$ 
& $\mathbf{1.14{\pm}0.20}$ 
& $\mathbf{0.91{\pm}0.03}$ 
& $\mathbf{1.03{\pm}0.11}$ 
& $\mathbf{0.94{\pm}0.02}$ 
& $\underline{1.08{\pm}0.17}$ \\
& MS-HVED 
& $\mathbf{0.86{\pm}0.04}$ 
& $\underline{1.16{\pm}0.20}$ 
& $\mathbf{0.91{\pm}0.03}$ 
& $\underline{1.04{\pm}0.13}$ 
& $\mathbf{0.94{\pm}0.02}$ 
& $\mathbf{1.07{\pm}0.16}$ \\
\bottomrule
\end{tabular}
\end{table*}

\subsection{Real-data evaluation}
\label{sec:res:real}

\subsubsection{Paired real cohort}

The $7$ paired real cases permitted comparison with a separately acquired HR pseudo-reference. As presented in \Cref{tab:real_comparison}, the FOV-weighted stack average achieved the highest PSNR, SSIM, and NCC. These metrics tend to favour smoother reconstructions with strong voxel-wise and structural agreement, and may therefore not fully reflect perceptual sharpness or preservation of fine anatomical detail. MS-HVED achieved the lowest perceptual loss and improved over SegResNet in every reported metric. However, the HR images were separate acquisitions rather than exact ground truth. We therefore retained these full-reference results and complemented them with expert ratings of the same seven cases in \Cref{sec:res:expert_review}.

\begin{table*}[pos=!tp]
\centering
\caption{\textbf{Full-reference evaluation on the $7$ paired real cases.} The independently acquired HR images were treated as pseudo-references. Values are mean{\scriptsize$\pm$}standard deviation. Best and second-best values computed in the present evaluation are bolded and underlined; published NiftyMIC values are included for context and excluded from this ranking.}
\label{tab:real_comparison}
\small
\setlength{\tabcolsep}{5pt}
\begin{tabular}{lcccc}
\toprule
\textbf{Method} & \textbf{PSNR}$\uparrow$ & \textbf{SSIM}$\uparrow$ & \textbf{NCC}$\uparrow$ & \textbf{Percept.}$\downarrow$ \\
\midrule
NiftyMIC~\cite{Ebner2019}
& 15.135{\scriptsize$\pm$2.90} 
& .374{\scriptsize$\pm$.043} 
& .748{\scriptsize$\pm$.043} 
& -- \\
NiftyMIC(reproduced)
& 18.82{\scriptsize$\pm$0.89}
& .604{\scriptsize$\pm$.115}
& \underline{.702}{\scriptsize$\pm$.069}
& .412{\scriptsize$\pm$.113} \\
ECLARE~\cite{eclare}
& 21.88{\scriptsize$\pm$1.05}
& .792{\scriptsize$\pm$.014}
& .622{\scriptsize$\pm$.051}
& .435{\scriptsize$\pm$.034} \\
TripleSR~\cite{triplesr}
& 21.56{\scriptsize$\pm$0.90}
& .771{\scriptsize$\pm$.014}
& .584{\scriptsize$\pm$.043}
& .498{\scriptsize$\pm$.025} \\
Naive mean              
& 22.18{\scriptsize$\pm$0.89} 
& .797{\scriptsize$\pm$.012} 
& .656{\scriptsize$\pm$.029} 
& .426{\scriptsize$\pm$.016}   \\
Naive + FOV             
& \textbf{23.35}{\scriptsize$\pm$1.09} 
& \textbf{.841}{\scriptsize$\pm$.009} 
& \textbf{.750}{\scriptsize$\pm$.035} 
& .352{\scriptsize$\pm$.022}   \\
SegResNet               
& 22.22{\scriptsize$\pm$1.08} 
& .813{\scriptsize$\pm$.013} 
& .652{\scriptsize$\pm$.065} 
& \underline{.277}{\scriptsize$\pm$.021}   \\
MS-HVED                 
& \underline{22.70}{\scriptsize$\pm$1.06} 
& \underline{.835}{\scriptsize$\pm$.009} 
& .697{\scriptsize$\pm$.054} 
& \textbf{.244}{\scriptsize$\pm$.016}   \\
\bottomrule
\end{tabular}
\end{table*}

\subsubsection{Reference-free clinical cohorts}

\begin{figure*}[!t]
\centering
\includegraphics[width=.8\textwidth]{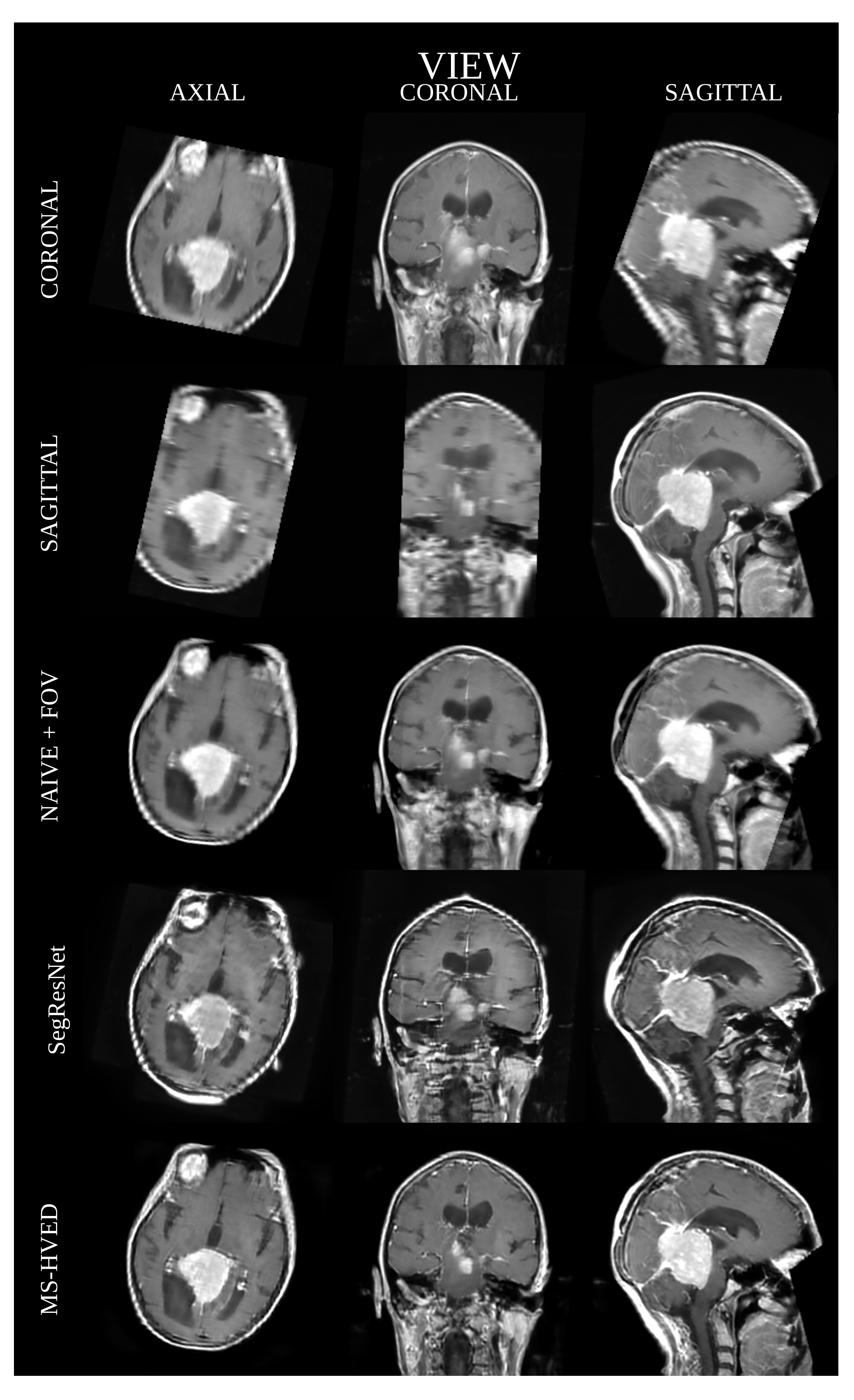}
\caption{\textbf{Qualitative meningioma reconstructions from real low-resolution stacks.} Examples show the available anisotropic clinical input stacks and the corresponding Naive+FOV, SegResNet, and MS-HVED reconstructions.}
\label{fig:meningioma_cases}
\end{figure*}

The meningioma examples in \Cref{fig:meningioma_cases} show reconstruction behaviour in the presence of tumour pathology and naturally missing orientations. Qualitative examples from the paired healthy and Nigerian cohorts are provided in the appendix (\Cref{fig:appendix_real_cases,fig:appendix_nigerian_cases}). As no matched HR acquisitions existed for the meningioma and Nigerian cohorts, conclusions for these data are based on the reader study rather than full-reference metrics.

\subsection{Expert-reader results}
\label{sec:res:expert_review}

Seven board-certified neuroradiologists evaluated 44 cases (30 Nigerian clinical cases, 7 meningioma cases, and 7 healthy-control cases), yielding 308 paired case--reader assessments. By cohort, these comprised 210 Nigerian, 49 meningioma, and 49 healthy-control assessments. By input condition, 182 assessments used every available stack, 42 omitted an available axial stack, and 84 omitted an available coronal stack. Pathology visibility was applicable only to the 7 meningioma cases and had 49 paired assessments.

\subsubsection{Results stratified by cohort}

As shown in \Cref{tab:reader_cohort}, MS-HVED received higher mean scores than SegResNet for every applicable criterion in the Nigerian and meningioma cohorts. All comparisons were significant with $p<.001$. For overall quality, MS-HVED was preferred in 187 of 210 Nigerian pairs and 47 of 49 meningioma pairs.

The paired real cohort showed a smaller and more criterion-dependent effect. MS-HVED received higher ratings for perceived SNR ($p<.001$), freedom from artefact ($p<.001$), anatomical consistency ($p=.004$), and overall quality ($p=.002$). Sharpness did not differ ($p=.145$), with SegResNet rated higher in 31 of 49 pairs.

\begin{table*}[!t]
\centering
\caption{\textbf{Reader scores stratified by real data cohort.} Each reader scored every case and both reconstructions. Scores are mean{\scriptsize$\pm$} standard deviation. Win ($W$) and Loss ($L$) count case reader pairs in which MS-HVED received a higher or lower score than SegResNet. $p$-values are from two-sided paired Wilcoxon signed-rank tests.}
\label{tab:reader_cohort}
\footnotesize
\setlength{\tabcolsep}{4pt}

\begin{tabular}{llrcccc}
\toprule
\textbf{Cohort} 
& \textbf{Criterion} 
& $n$ & \textbf{MS-HVED} 
& \textbf{SegResNet} 
& $W/L$ & $p$ \\
\midrule
\multirow{5}{*}{Nigerian}
& SNR 
& 210 & $5.67\pm2.08$ & $4.08\pm2.14$ & 184/26 & $<.001$ \\
& Sharpness & 210 & $5.32\pm2.07$ & $4.26\pm2.14$ & 159/51 & $<.001$ \\
& Artefact & 210 & $5.15\pm1.93$ & $3.30\pm1.85$ & 199/11 & $<.001$ \\
& Consistency & 210 & $5.47\pm1.92$ & $3.75\pm1.87$ & 186/24 & $<.001$ \\
& Overall quality & 210 & $5.27\pm1.88$ & $3.58\pm1.82$ & 187/23 & $<.001$ \\
\midrule
\multirow{6}{*}{Meningioma}
& SNR & 49 & $6.39\pm1.86$ & $4.37\pm2.00$ & 47/2 & $<.001$ \\
& Sharpness & 49 & $5.90\pm2.02$ & $4.37\pm1.86$ & 37/12 & $<.001$ \\
& Artefact & 49 & $6.16\pm1.68$ & $4.27\pm1.85$ & 48/1 & $<.001$ \\
& Consistency & 49 & $6.24\pm1.83$ & $4.43\pm1.80$ & 46/3 & $<.001$ \\
& Pathology visibility & 49 & $6.24\pm1.97$ & $4.76\pm1.91$ & 44/5 & $<.001$ \\
& Overall quality & 49 & $6.18\pm1.68$ & $4.14\pm1.65$ & 47/2 & $<.001$ \\
\midrule
\multirow{5}{*}{Paired real}
& SNR & 49 & $7.31\pm1.62$ & $6.47\pm1.67$ & 38/11 & $<.001$ \\
& Sharpness & 49 & $6.47\pm1.60$ & $6.71\pm1.66$ & 18/31 & $.145$ \\
& Artefact & 49 & $6.84\pm1.55$ & $5.84\pm1.57$ & 39/10 & $<.001$ \\
& Consistency & 49 & $6.76\pm1.44$ & $6.08\pm1.50$ & 34/15 & $.004$ \\
& Overall quality & 49 & $6.88\pm1.30$ & $6.20\pm1.46$ & 34/15 & $.002$ \\
\bottomrule
\end{tabular}
\end{table*}

\subsubsection{Results stratified by input availability}

We present results based on input availability in \Cref{tab:reader_input}. The overall-quality analysis of the full-input group favoured MS-HVED. Within this group, MS-HVED was also rated higher for SNR, freedom from artefact, anatomical consistency, pathology visibility, and sharpness (all $p<.001$).

The result persisted when an available orientation was withheld from both methods. Overall quality favoured MS-HVED in 40 of 42 axial-omitted pairs and 76 of 84 coronal-omitted pairs (both $p<.001$), all other applicable criteria in these two groups also had $p<.001$.

\begin{table*}[!t]
\centering
\caption{\textbf{Reader scores stratified by reconstruction input condition.} Full input denotes use of every real data stack available for a case. In the omitted groups, the named acquired stack was shown to readers but withheld from both methods during inference. Values and tests are as defined as in \Cref{tab:reader_cohort}.}
\label{tab:reader_input}
\footnotesize
\setlength{\tabcolsep}{4pt}
\begin{tabular}{llrcccc}
\toprule
\textbf{Input group} & \textbf{Criterion} & $n$ & \textbf{MS-HVED} & \textbf{SegResNet} & $W/L$ & $p$ \\
\midrule
\multirow{6}{*}{Full input}
& SNR & 182 & $6.25\pm1.97$ & $4.67\pm2.21$ & 161/21 & $<.001$ \\
& Sharpness & 182 & $5.66\pm1.99$ & $5.09\pm2.15$ & 100/82 & $<.001$ \\
& Artefact & 182 & $5.84\pm1.88$ & $4.27\pm2.03$ & 166/16 & $<.001$ \\
& Consistency & 182 & $5.97\pm1.83$ & $4.62\pm1.93$ & 151/31 & $<.001$ \\
& Pathology visibility & 49 & $6.24\pm1.97$ & $4.76\pm1.91$ & 44/5 & $<.001$ \\
& Overall quality & 182 & $5.93\pm1.76$ & $4.49\pm1.94$ & 152/30 & $<.001$ \\
\midrule
\multirow{5}{*}{Axial omitted}
& SNR & 42 & $5.36\pm2.25$ & $3.74\pm2.24$ & 35/7 & $<.001$ \\
& Sharpness & 42 & $5.12\pm2.27$ & $3.21\pm1.99$ & 41/1 & $<.001$ \\
& Artefact & 42 & $4.76\pm2.16$ & $2.67\pm1.80$ & 41/1 & $<.001$ \\
& Consistency & 42 & $5.07\pm2.26$ & $3.07\pm1.81$ & 39/3 & $<.001$ \\
& Overall quality & 42 & $4.88\pm2.23$ & $2.81\pm1.76$ & 40/2 & $<.001$ \\
\midrule
\multirow{5}{*}{Coronal omitted}
& SNR & 84 & $5.93\pm2.10$ & $4.54\pm2.18$ & 73/11 & $<.001$ \\
& Sharpness & 84 & $5.69\pm2.01$ & $4.49\pm2.13$ & 73/11 & $<.001$ \\
& Artefact & 84 & $5.42\pm1.86$ & $3.57\pm1.85$ & 79/5 & $<.001$ \\
& Consistency & 84 & $5.79\pm1.78$ & $3.95\pm1.93$ & 76/8 & $<.001$ \\
& Overall quality & 84 & $5.50\pm1.80$ & $3.86\pm1.87$ & 76/8 & $<.001$ \\
\bottomrule
\end{tabular}
\end{table*}

\subsubsection{Scale-invariant preference and inter-rater variability}

\begin{figure*}[!t]
\centering
\includegraphics[width=\textwidth]{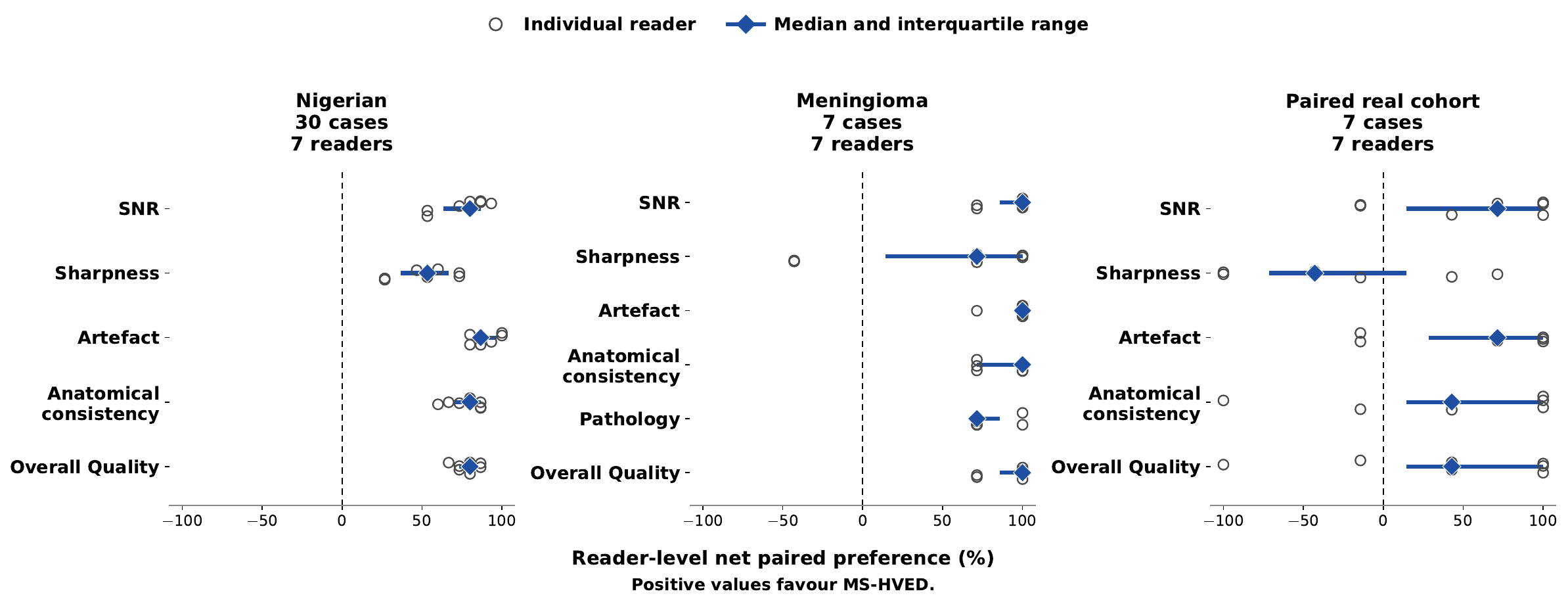}
\caption{\textbf{Scale-invariant reader preference by real data cohort.} For every reader and criterion, net paired preference is $100(W-L)/(W+L)$, where $W$ and $L$ are case-level wins and losses for MS-HVED relative to SegResNet. Circles show individual readers, diamonds and horizontal lines show the median and interquartile range. Positive values favour MS-HVED and negative values favour SegResNet. Because the measure uses only the within-pair ordering, it is unchanged by reader-specific shifts or positive rescaling of scores.}
\label{fig:reader_scale_invariant}
\end{figure*}

The reader-level net paired-preference analysis in \Cref{fig:reader_scale_invariant} removes differences in how readers used the numerical scale. For overall quality, every reader had positive net preference for MS-HVED in the Nigerian and meningioma cohorts. In the paired real cohort, five readers favoured MS-HVED and two favoured SegResNet, explaining the wider reader-level dispersion despite the positive pooled result. The same pattern was present for SNR, freedom from artefact, and anatomical consistency. Sharpness was the least consistent criterion, where SegResNet was favoured more than MS-HVED due to its overly sharp reconstructions. Aggregate win/loss summaries and radar charts are provided in the supplementary material (\Cref{fig:supp_reader_preference,fig:supp_reader_radar_full,fig:supp_reader_radar_missing}).

Inter-rater agreement was criterion dependent (\Cref{tab:reader_agreement}). Median pairwise quadratic-weighted $\kappa$ was highest for overall quality (0.46) and lowest for pathology visibility (0.27), with the latter based only on the seven meningioma cases. Across criteria, 38--47\% of reader pairs were within one score point. These results indicate appreciable variation in the absolute scores assigned by different readers.

\begin{table*}[!t]
\centering
\caption{\textbf{Inter-rater variability.} Values were calculated across shared case--method units. Quadratic-weighted $\kappa$ is reported as the median and range across reader pairs. Within-one-point agreement and mean absolute difference (MAD) are averaged across reader pairs.}
\label{tab:reader_agreement}
\small
\setlength{\tabcolsep}{7pt}
\begin{tabular}{lccc}
\toprule
\textbf{Criterion} & \textbf{Weighted $\kappa$} & \textbf{Within one point} & \textbf{MAD} \\
\midrule
SNR & $0.34\;(0.05\text{--}0.58)$ & $0.39$ & $2.31$ \\
Sharpness & $0.30\;(0.06\text{--}0.59)$ & $0.38$ & $2.24$ \\
Freedom from artefact & $0.39\;(0.08\text{--}0.73)$ & $0.44$ & $2.01$ \\
Anatomical consistency & $0.37\;(0.08\text{--}0.61)$ & $0.43$ & $1.96$ \\
Pathology visibility & $0.27\;(0.06\text{--}0.58)$ & $0.38$ & $2.20$ \\
Overall quality & $0.46\;(0.08\text{--}0.68)$ & $0.47$ & $1.86$ \\
\bottomrule
\end{tabular}
\end{table*}

\begin{figure*}[pos=p]
    \centering
    \includegraphics[width=0.9\textwidth]{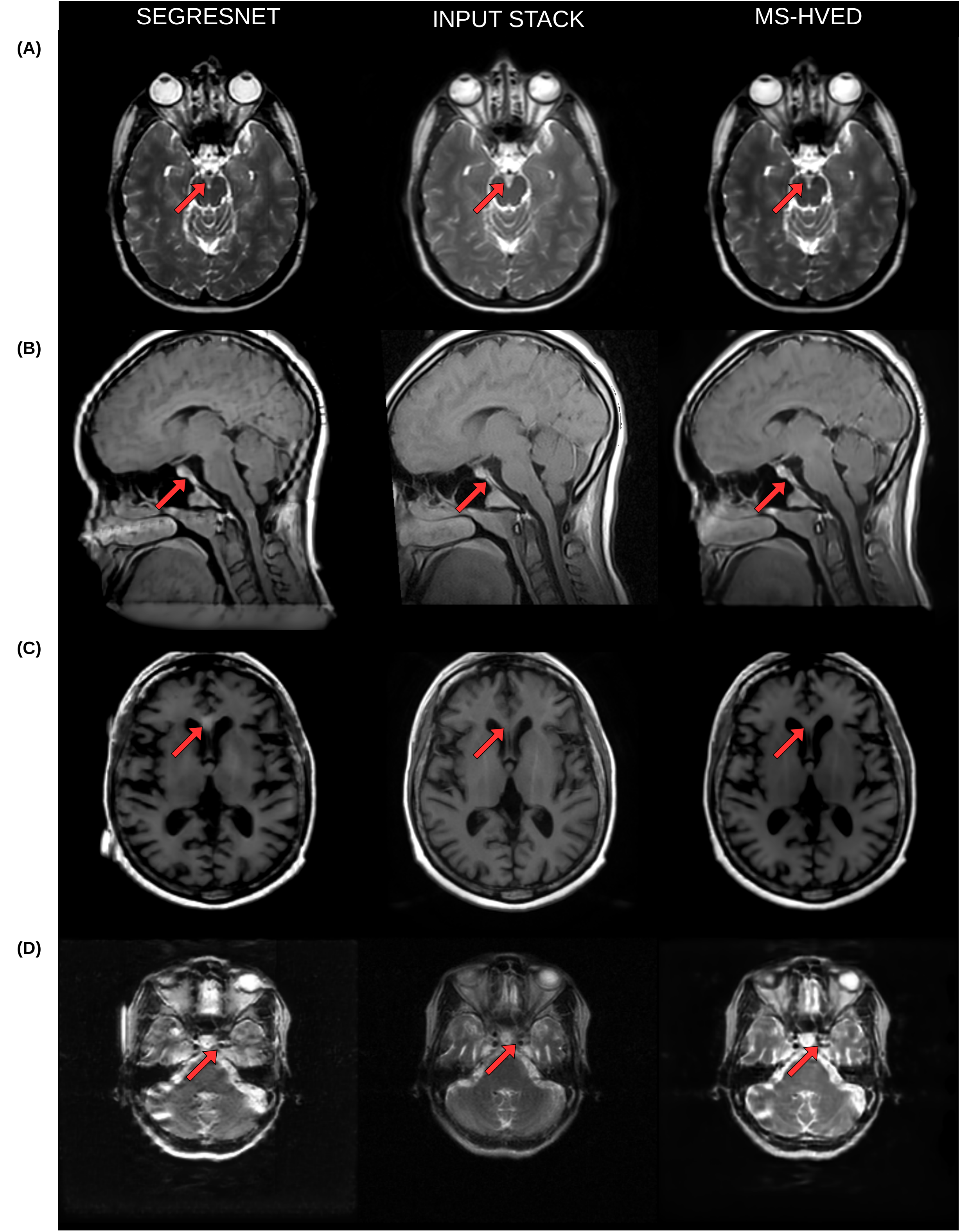}
    \caption{\textbf{Representative anatomical observations identified during expert review.}
    Examples show local reconstruction errors that were not necessarily reflected by global image-quality metrics.
    (a) Healthy-control case showing reduced conspicuity of the interpeduncular cistern in both reconstructions.
    (b) Nigerian case showing reduced distinction between the pituitary gland and clivus in the SegResNet reconstruction.
    (c) Nigerian case showing increased signal intensity in the genu of the corpus callosum in the SegResNet reconstruction.
    (d) Nigerian $0.3$\,T case showing reduced separation of the left ICA flow void from adjacent structures in both reconstructions, more prominently with SegResNet.
    Arrows indicate the regions discussed by the expert readers.}
    \label{fig:reader_observations}
\end{figure*}
\section{Discussion}
\label{sec:discussion}

This study demonstrates that MS-HVED can reconstruct isotropic brain MRI from variable subsets of anisotropic multi-orientation stacks while accounting for the spatial extent of each acquisition. Across complementary evaluations, the method achieved competitive or leading reconstruction fidelity, preserved downstream tissue-segmentation performance, and was preferred to SegResNet by expert neuroradiologists when evaluating real clinical data.

\subsection{Acquisition-aware fusion improves robustness to clinical geometry}

Geometric ablation provides the clearest evidence for the importance of acquisition-aware modelling. Introducing oblique and partial-FOV simulation during training increased PSNR by $4.46$--$6.06$~dB compared with training without realistic acquisition geometry. Incorporating FOV masks produced further improvements, particularly in SSIM and perceptual loss.

These components address related but distinct sources of variation. The domain-randomisation pipeline exposes the model to rotated, incomplete, and partially overlapping acquisitions during training, whereas the FOV masks determine where each orientation should contribute during fusion. This allows MS-HVED to reconstruct over the union of available acquisitions without treating resampled background or unsupported voxels as valid measurements.

\subsection{MS-HVED supports variable orientations, contrasts, 
reconstruction grids, 
and stochastic inference}

A key benefit of MS-HVED is that one trained model can operate on different subsets of axial, coronal and sagittal stacks. The model maintained similar performance across \textsc{Ax+Cor}, \textsc{Ax+Sag} and \textsc{Cor+Sag} configurations without requiring a separate network for each input combination, unlike fixed-input supervised baselines. This flexibility is important for retrospective and clinical datasets, where stacks may be absent because of protocol differences, incomplete examinations, motion corruption or acquisition failure.

The sequence-agnostic experiment demonstrated a related trade-off. A single model jointly trained on IXI-T1, IXI-T2, and \KCHDAT remained competitive with dataset-specific models and improved several PSNR, MAE, and perceptual results, although it did not achieve the best result for every metric. 

Additionally, since MS-HVED is fully convolutional, its inference is not tied to a fixed grid. A model trained using 1mm isotropic targets can therefore be evaluated on a finer reconstruction grid, e.g. 0.5mm. We demonstrate this resolution flexibility across acquisitions with different native resolutions and field strengths in \Cref{appendix:multi_res}.

Lastly, the variational formulation of MS-HVED provides an additional capability beyond the deterministic reconstructions evaluated. At inference, the fused multi-scale posterior can be decoded at its mean, as used for all reported experiments, or sampled repeatedly to produce stochastic multi-sample reconstructions. Repeated samples can expose regions in which the reconstruction is sensitive to latent uncertainty. This is particularly relevant where one or more orientations are missing or have limited coverage. Establishing such dispersion as a calibrated measure of reconstruction uncertainty will require dedicated validation, nevertheless, multi-sample inference follows directly from the fused posterior and does not require a change to the trained architecture.

\subsection{Full-reference metrics do not capture all reconstruction artefacts}

The experiments highlight the limitations of conventional full-reference metrics for MRI super-resolution. MAE, PSNR and SSIM are useful for controlled synthetic evaluation, but they aggregate errors over the reconstructed volume and may hide localised boundary or wedge-shaped artefacts. This helps explain why SegResNet was competitive on several simulated benchmarks, while expert readers identified poorer image quality. These localised errors include altered signal intensity, loss of anatomical separation, and reduced conspicuity of small structures.

\subsection{Expert assessment provides evidence on real clinical image quality}

The blinded reader study extends the evaluation beyond simulated degradation models and imperfect pseudo-references. Seven expert neuroradiologists assessed 44 cases from the paired healthy, Nigerian, and meningioma cohorts. MS-HVED received higher overall-quality ratings than SegResNet when all available stacks were used and retained its advantage when axial or coronal inputs were withheld. Improvements were also observed for artefact, SNR, and anatomical consistency, while differences in sharpness were less pronounced. This pattern suggests that the preference for MS-HVED was not driven solely by apparent sharpness, but also by cleaner and more anatomically coherent reconstructions.

The accompanying reader comments provide further insight into the types of residual errors present in both methods. \Cref{fig:reader_observations} shows examples of these errors with reader comments. In one healthy-control case, the interpeduncular cistern appeared abnormally dark in both reconstructions, indicating a failure shared by the two learned approaches. In a meningioma case with no sagittal input, the spinal cord was poorly delineated in both reconstructions, although SegResNet showed more pronounced artefacts. Readers also identified several local anatomical inconsistencies in the Nigerian cohort: increased signal within the genu of the corpus callosum in a SegResNet reconstruction, conflation of the left ICA flow void with adjacent structures in both methods but more prominently with SegResNet, and reduced separation between the pituitary gland and clivus in another SegResNet case. These observations complement the quantitative ratings by showing that residual reconstruction errors can be spatially localised and anatomically specific, and may therefore have limited influence on global image-similarity metrics.

\subsection{Limitations and future work}

Several limitations remain. First, MS-HVED assumes that the available stacks have been rigidly registered and resampled to a common space before fusion. Residual misregistration can introduce inconsistent anatomical evidence that the fusion mechanism cannot fully correct, and registration error is not explicitly simulated during training. Second, although the reader study included multiple acquisition sources, the clinical meningioma cohort contained only seven cases and therefore represented a limited range of tumour appearances and imaging protocols. Larger multi-institutional studies are needed to assess robustness across scanners, field strengths, contrasts, pathologies and acquisition quality, and to determine whether improvements in reconstructed image quality translate into improved diagnostic interpretation.

The reader comments also indicate that acquisition-domain shift warrants further investigation. Row (D) of \Cref{fig:reader_observations} shows a Nigerian case acquired at $0.3$,T, both methods showed reduced separation of the left ICA flow void from adjacent structures, with the artefact more pronounced for SegResNet. Since the training cohorts consisted predominantly of $1.5$,T and $3$,T acquisitions, differences in SNR and image characteristics at low field may have contributed to this failure. Future work should therefore include dedicated low-field training and evaluation data. 

Finally, the discrepancy between competitive simulated metrics and expert observations of local anatomical errors highlights the need for SRR metrics that are sensitive to boundary artefacts, altered signal, anatomical separation and pathology preservation. The stochastic posterior available in MS-HVED may also provide useful uncertainty information, but its calibration and clinical relevance require dedicated validation.

\afterpage{\clearpage}
\section{Conclusion}
\label{sec:conclusion}
MS-HVED provides a flexible variational framework for reconstructing isotropic brain MRI from the available subset of anisotropic axial, coronal, and sagittal image stacks. Its combination of oblique partial-FOV simulation and spatially aware Product-of-Gaussians fusion allows the model to use valid information from each orientation without assuming identical extent or requiring a separately trained network for every input configuration.

Across synthetic and real (paired and reference-free) clinical evaluations, MS-HVED achieved competitive reconstruction fidelity, preserved downstream tissue segmentation performance, and received higher expert ratings than SegResNet for overall image quality. The reader preference persisted across the Nigerian, meningioma, and paired healthy cohorts and when an axial or coronal input was deliberately omitted. The variational posterior additionally permits stochastic multi-sample inference, providing a basis for future evaluation of reconstruction stability and uncertainty.

These results support MS-HVED as a robust approach for heterogeneous multi-stack brain MRI. Future multi-institutional evaluation will extend the range of datasets and pathologies, address registration-related failures, and determine whether the observed improvements translate into greater diagnostic accuracy and clinical utility.

\section*{Declaration of interest}

T. Vercauteren and J. Shapey are co-founders and shareholders of Hypervision Surgical, which has no interest in the present work.
The authors declare that they have no other known financial interests or personal relationships that could have appeared to influence the work reported in this paper.

\section*{Funding}

This work was supported by King's College London through the King's Doctoral College Africa Studentship. LG-FM acknowledges funding from the EPSRC Centre for Doctoral Training in Smart Medical Imaging  [grant number EP/S022104/1].

\section*{Declaration of generative AI and AI-assisted technologies in the manuscript preparation process}
During the preparation of this work the authors used OpenAI's ChatGPT (5.6 Sol) to assist with language editing, restructuring of author-written text, and improving clarity and concision. After using this tool, the authors reviewed and edited the content as needed and take full responsibility for the content of the published article.


\bibliographystyle{bibstyle}
\bibliography{references}

\FloatBarrier
\clearpage

\onecolumn
\appendix
\setcounter{figure}{0}
\setcounter{table}{0}
\renewcommand{\thefigure}{A.\arabic{figure}}
\renewcommand{\thetable}{A.\arabic{table}}
\renewcommand{\theHfigure}{A.\arabic{figure}}
\renewcommand{\theHtable}{A.\arabic{table}}

\section{Supporting information}
\label{appendix}

\subsection{Expert review interface}
\label{appendix:app}

\begin{figure}[pos=!hbt]
\centering
\includegraphics[width=\textwidth]{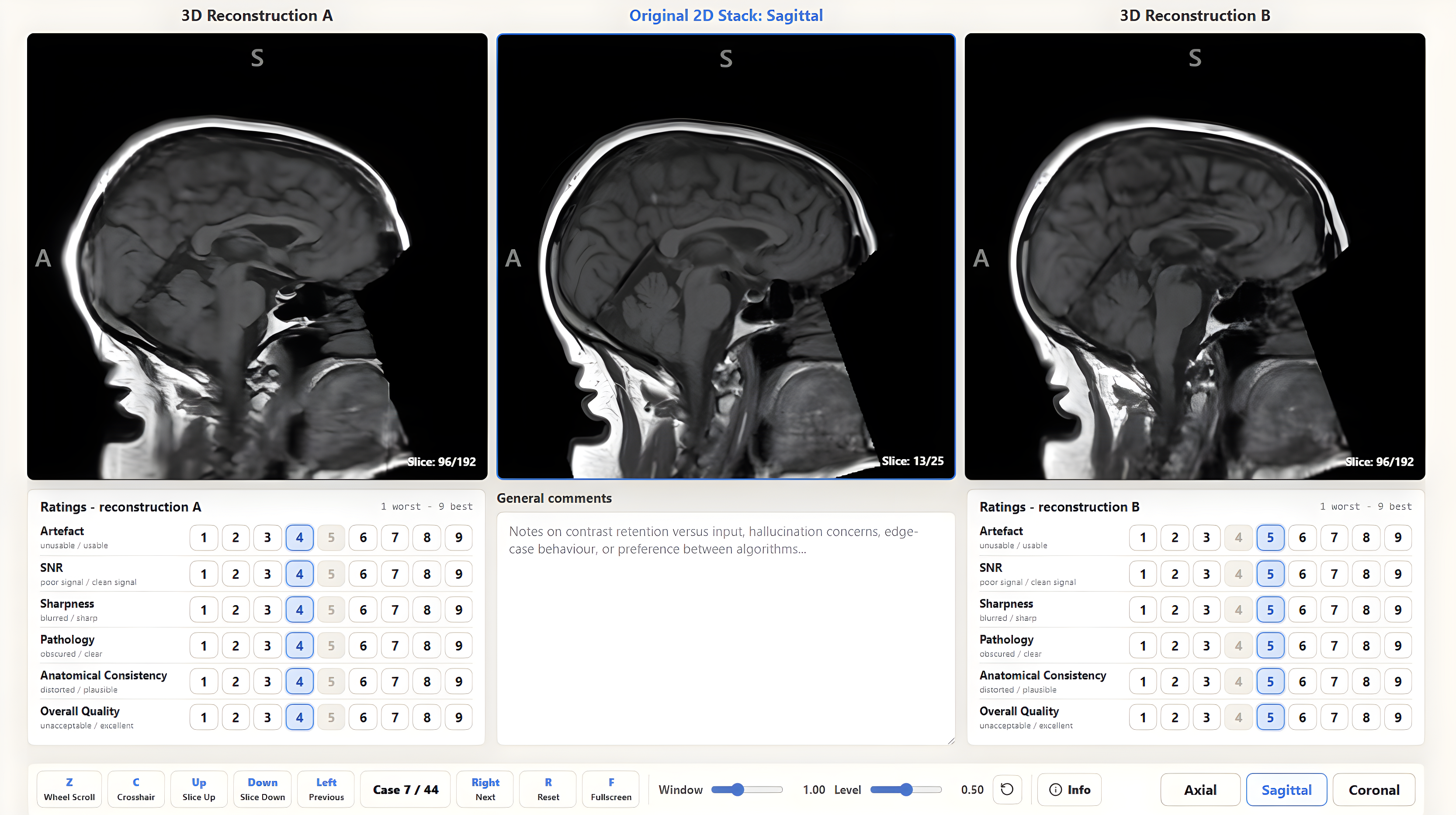}
\caption{%
\textbf{Expert review interface.}
Screenshot of the blinded multi-planar review interface used for reference-free clinical evaluation. The interface displays registered LR inputs and anonymised reconstructions side by side, allowing expert raters to inspect axial, coronal and sagittal views before assigning image-quality scores.}
\label{fig:reviewer_ui}
\end{figure}

For the reference-free clinical evaluation, expert raters reviewed reconstructions using a dedicated multi-planar visualisation interface as shown in \Cref{fig:reviewer_ui}. The interface was designed to support direct comparison between reconstruction methods while keeping method identities hidden from the reviewers. For each case, the available registered LR stacks were shown alongside the reconstructed isotropic volumes, allowing raters to inspect whether each reconstruction preserved the anatomical content visible in the input acquisitions.

\subsection{Supplementary qualitative results}
\label{appendix:qr}

\begin{figure*}[pos=!hbt]
    \centering
    \includegraphics[width=0.45\textwidth]{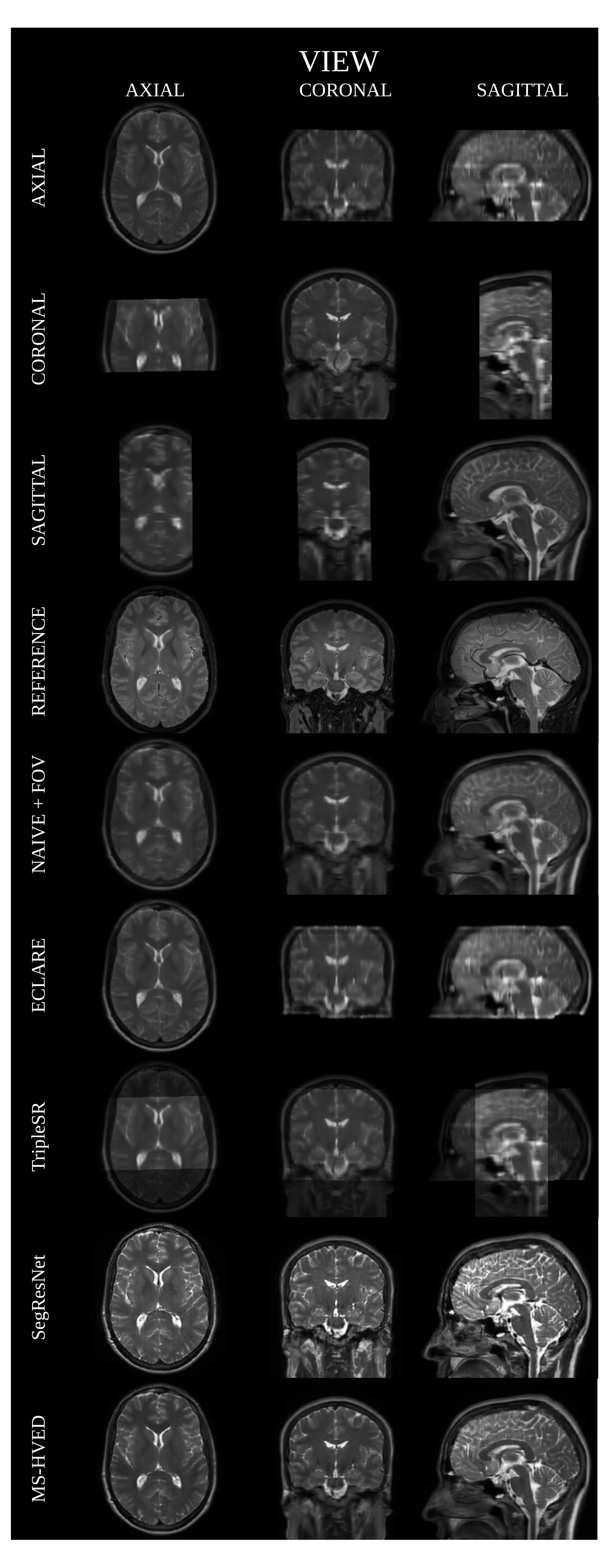}
    \caption{
    \textbf{Qualitative comparison on the healthy cohort (real data).} 
    Reconstructions are shown for a representative case from the paired healthy-control cohort \cite{Ebner2019}. Available low resolution (LR) stacks are shown in their inplane and through-plane orientations. Naive+FOV provide non-learned reference baselines obtained by averaging the registered LR stacks with FOV weighting. The separately acquired HR image is shown as a pseudo-reference rather than an absolute ground truth because of contrast differences between the LR and HR acquisitions.}
    \label{fig:appendix_real_cases}
\end{figure*}

\begin{figure}[pos=!btp]
    \centering
    \includegraphics[width=0.7\textwidth]{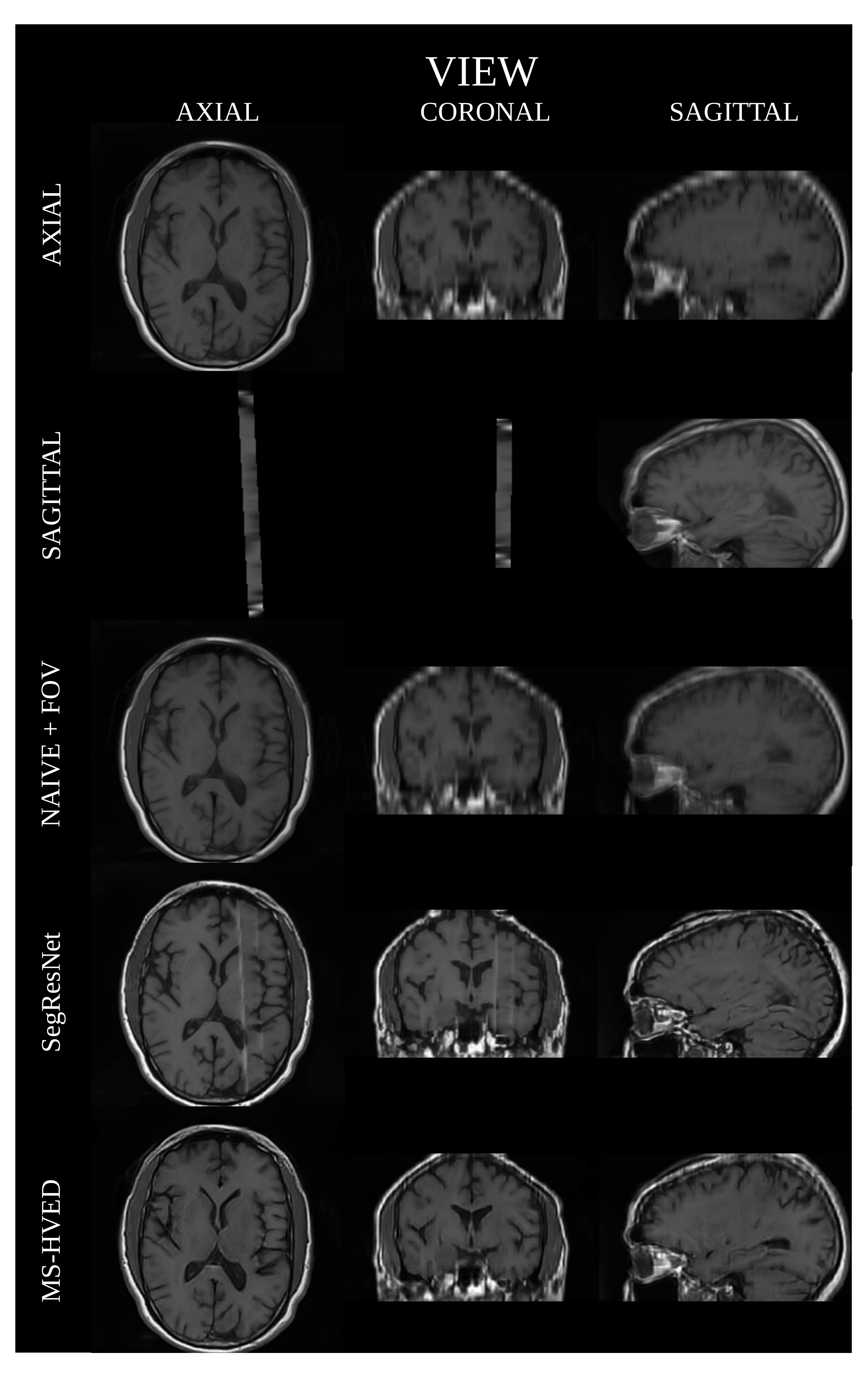}
    \caption{\textbf{Qualitative comparison on the Nigerian cohort (real data).}
    Reconstructions are shown for a representative case from the Nigerian clinical cohort \cite{wogu2025nigerian,nigerianbrainlife2024}. Available low resolution (LR) stacks are shown in their inplane and through-plane orientations. Naive+FOV provide non-learned reference baselines obtained by averaging the registered LR stacks with FOV weighting. SegResNet and MS-HVED correspond to the two strongest learned models and are used for the primary expert comparison.}
    \label{fig:appendix_nigerian_cases}
\end{figure}

\begin{figure}[pos=!btp]
    \centering
    \includegraphics[width=\textwidth]{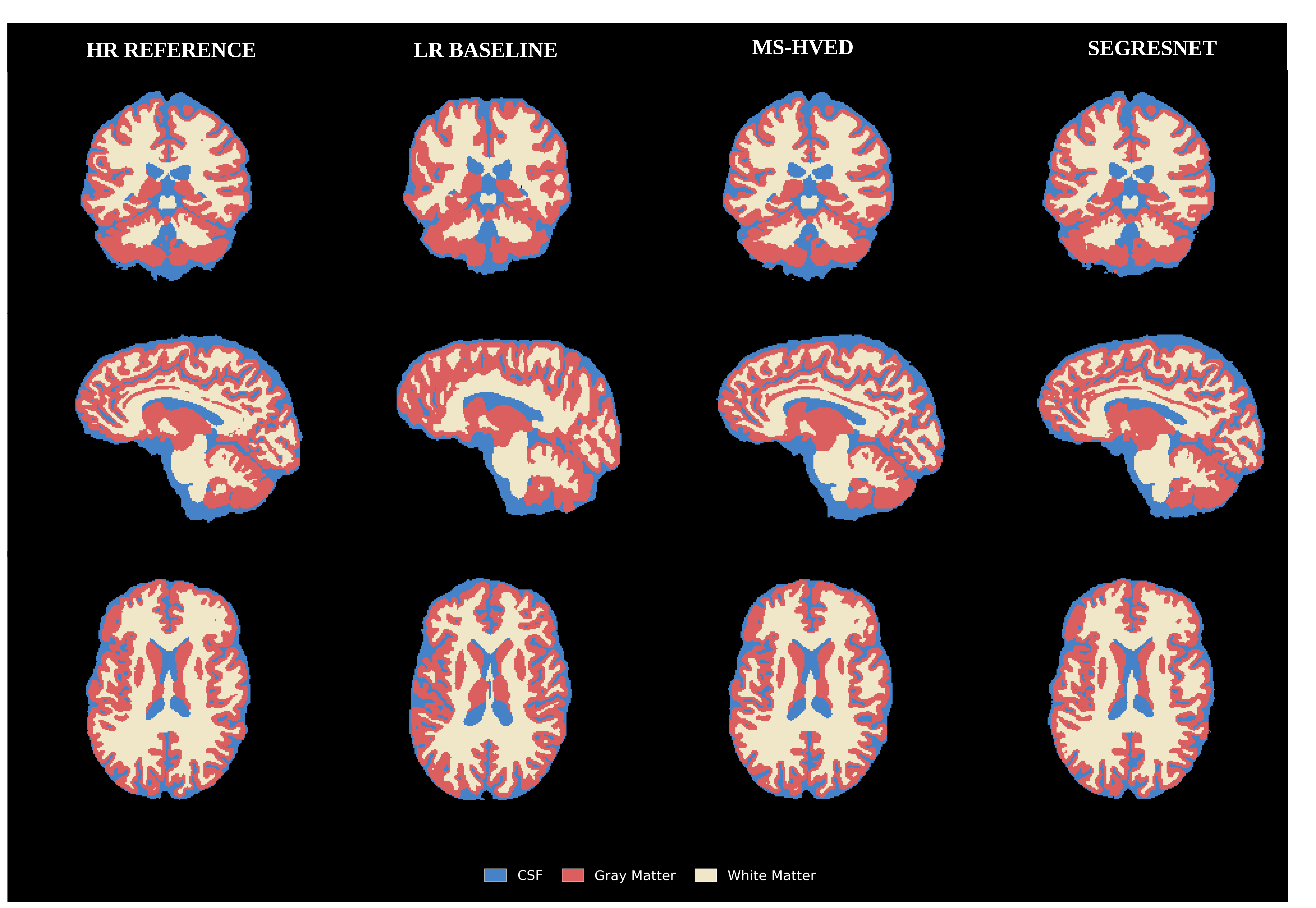}
    \caption{\textbf{Qualitative downstream brain-tissue segmentation comparison.} A representative SynthSeg~\cite{SynthSeg} segmentation from an IXI-T1 test case is shown for the LR baseline, SegResNet reconstruction and MS-HVED reconstruction, together with the HR reference segmentation. } 
    \label{fig:downstream_seg}
\end{figure}

We provide additional qualitative comparisons to complement the quantitative results reported in the main text. \Cref{fig:appendix_real_cases} and \Cref{fig:appendix_nigerian_cases} show representative reconstructions from the paired healthy-control and Nigerian clinical cohorts, respectively, while \Cref{fig:downstream_seg} shows downstream tissue segmentations.

\FloatBarrier
\subsection{Supplementary reader-study figures}
\label{appendix:reader_figures}

\begin{figure}[pos=!tbhp]
\centering
\includegraphics[width=.92\textwidth]{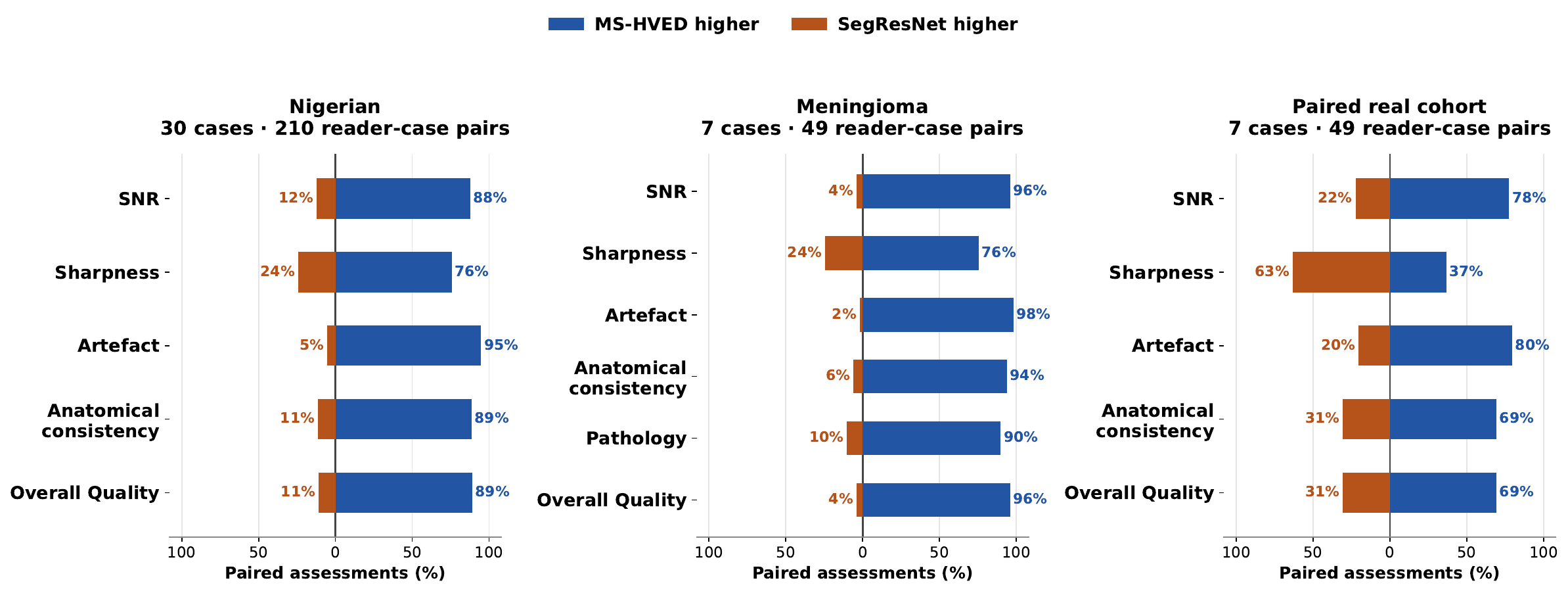}
\caption{\textbf{Aggregate paired preference by real data cohort.} Bars show the proportions of paired case--reader assessments in which MS-HVED received a higher score or a lower score than SegResNet. Comparisons are made within the same reader and case, so they do not depend on differences in the absolute score ranges used by different readers.}
\label{fig:supp_reader_preference}
\end{figure}

\begin{figure}[pos=!tbhp]
\centering
\includegraphics[width=.4\textwidth]{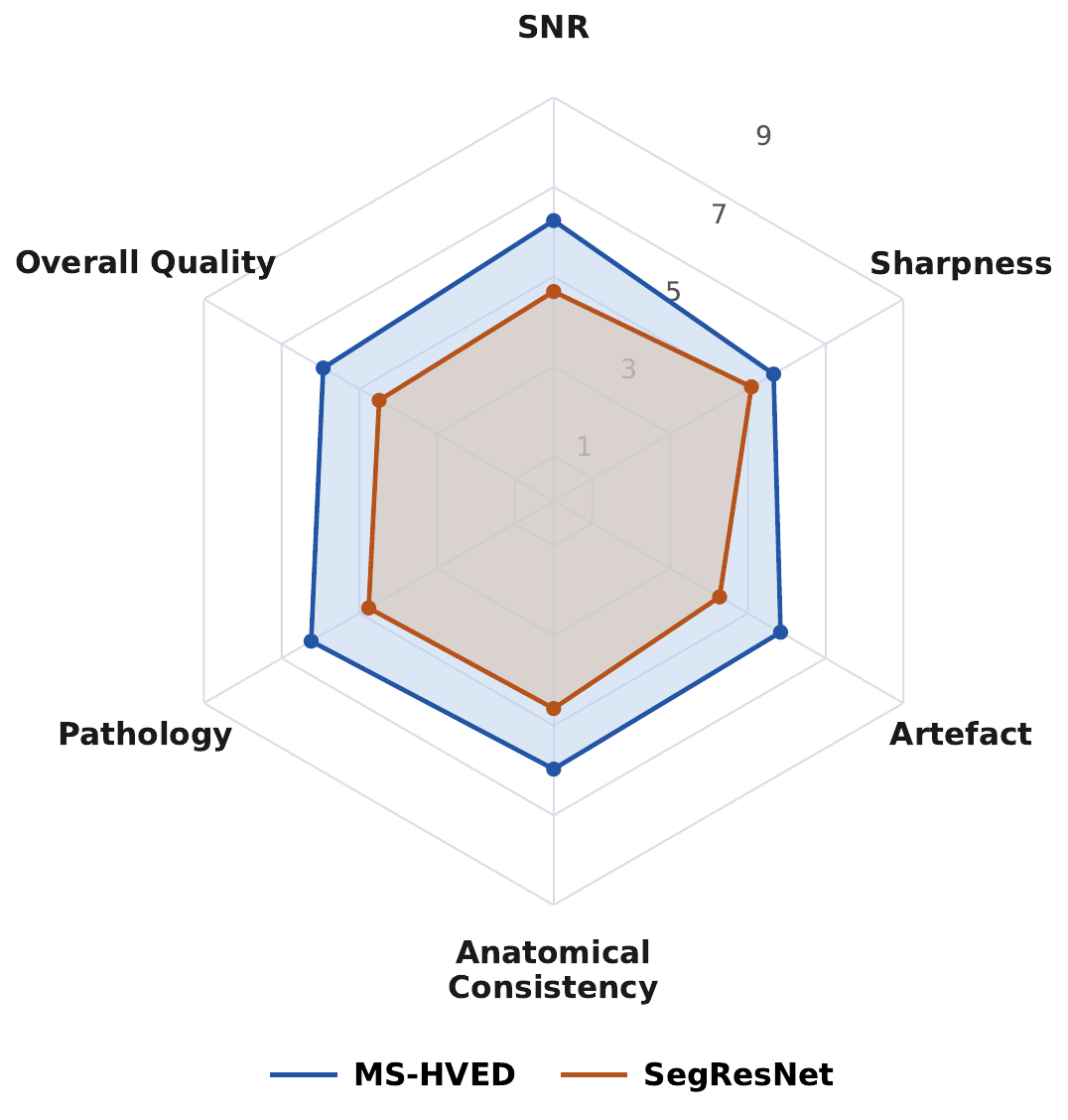}
\caption{\textbf{Mean reader scores for full-input reconstructions from real data.} The radar plot is retained as a compact supplementary overview while the box plots in the main text show the underlying distributions. Pathology visibility was scored only for meningioma cases.}
\label{fig:supp_reader_radar_full}
\end{figure}

\begin{figure}[pos=!tbhp]
\centering
\includegraphics[width=.4\textwidth]{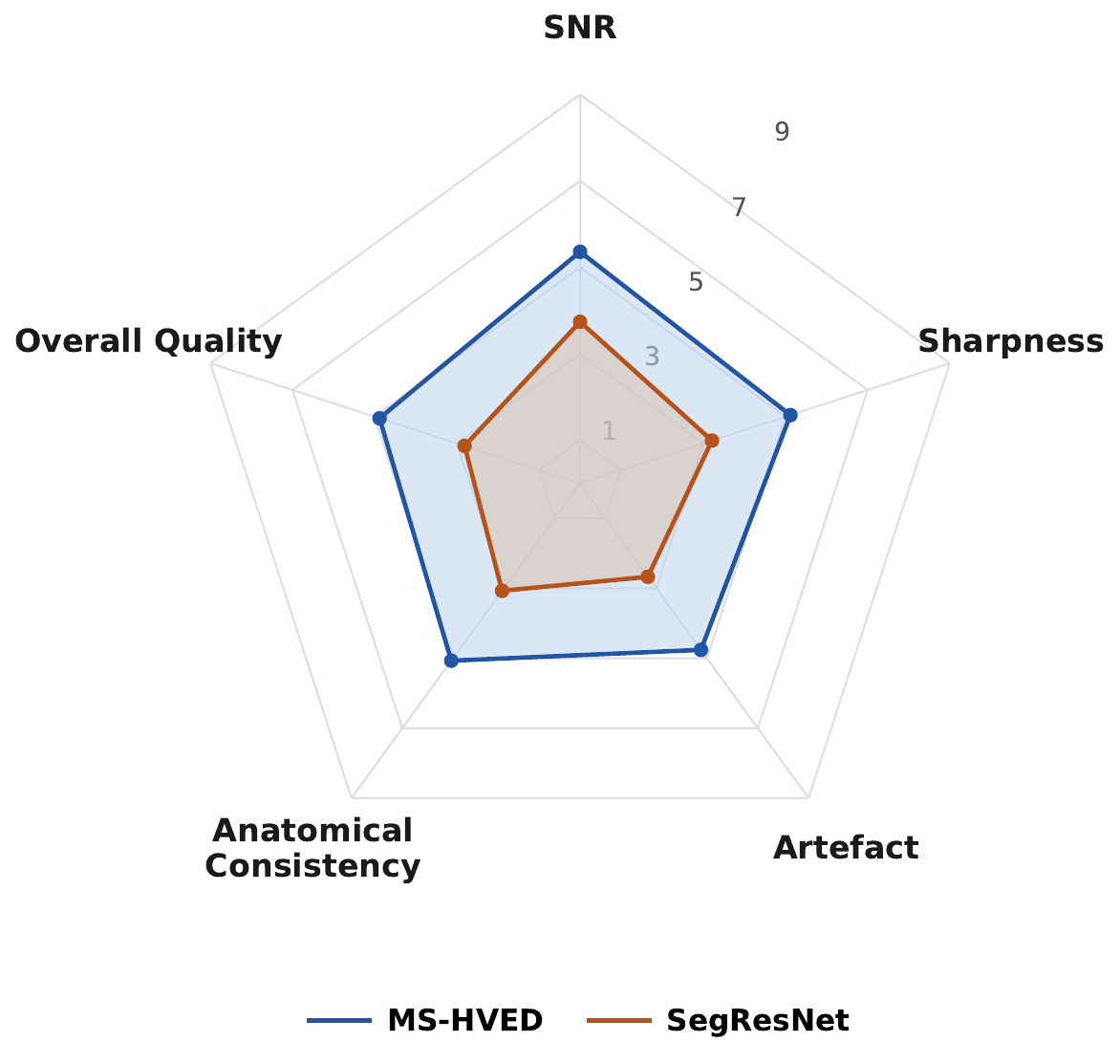}
\quad
\includegraphics[width=.4\textwidth]{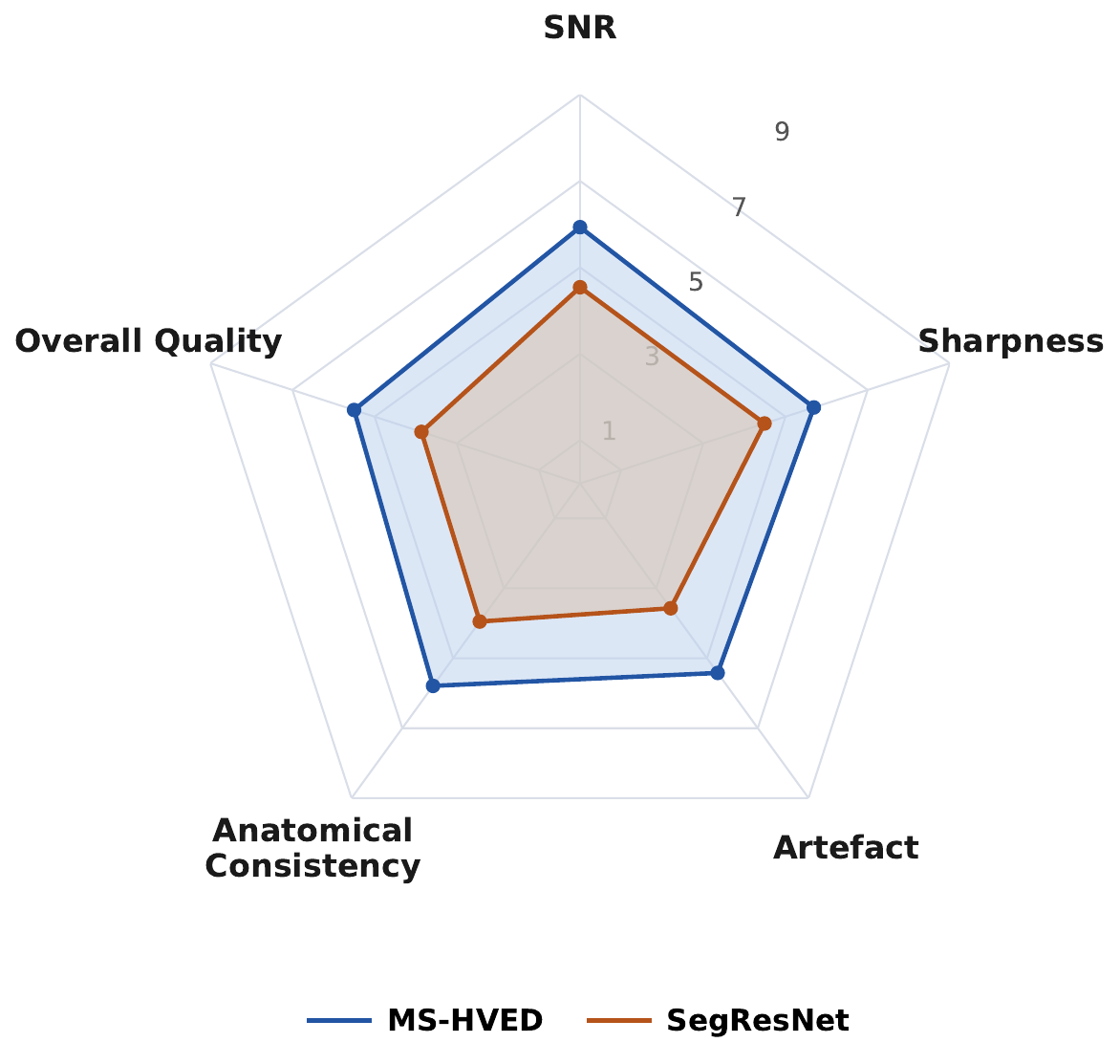}
\caption{\textbf{Mean reader scores for deliberately omitted inputs (real data).} The axial-omitted group is shown on the left and the coronal-omitted group on the right. These plots summarise the same data whose distributions are presented as box plots in the main text.}
\label{fig:supp_reader_radar_missing}
\end{figure}

Additional summaries of the expert-reader study are provided in \Cref{fig:supp_reader_preference,fig:supp_reader_radar_full,fig:supp_reader_radar_missing}. \Cref{fig:supp_reader_preference} reports paired method preferences within the same case and reader, reducing the influence of differences in how individual readers use the rating scale. The radar plots provide complementary summaries of the mean criterion-specific scores for full-input and deliberately omitted-input reconstructions.

\subsection{Examples of orientation-specific FOV coverage}
\label{appendix:sup}

\begin{figure}[pos=htbp]
    \centering
    \includegraphics[width=\textwidth]{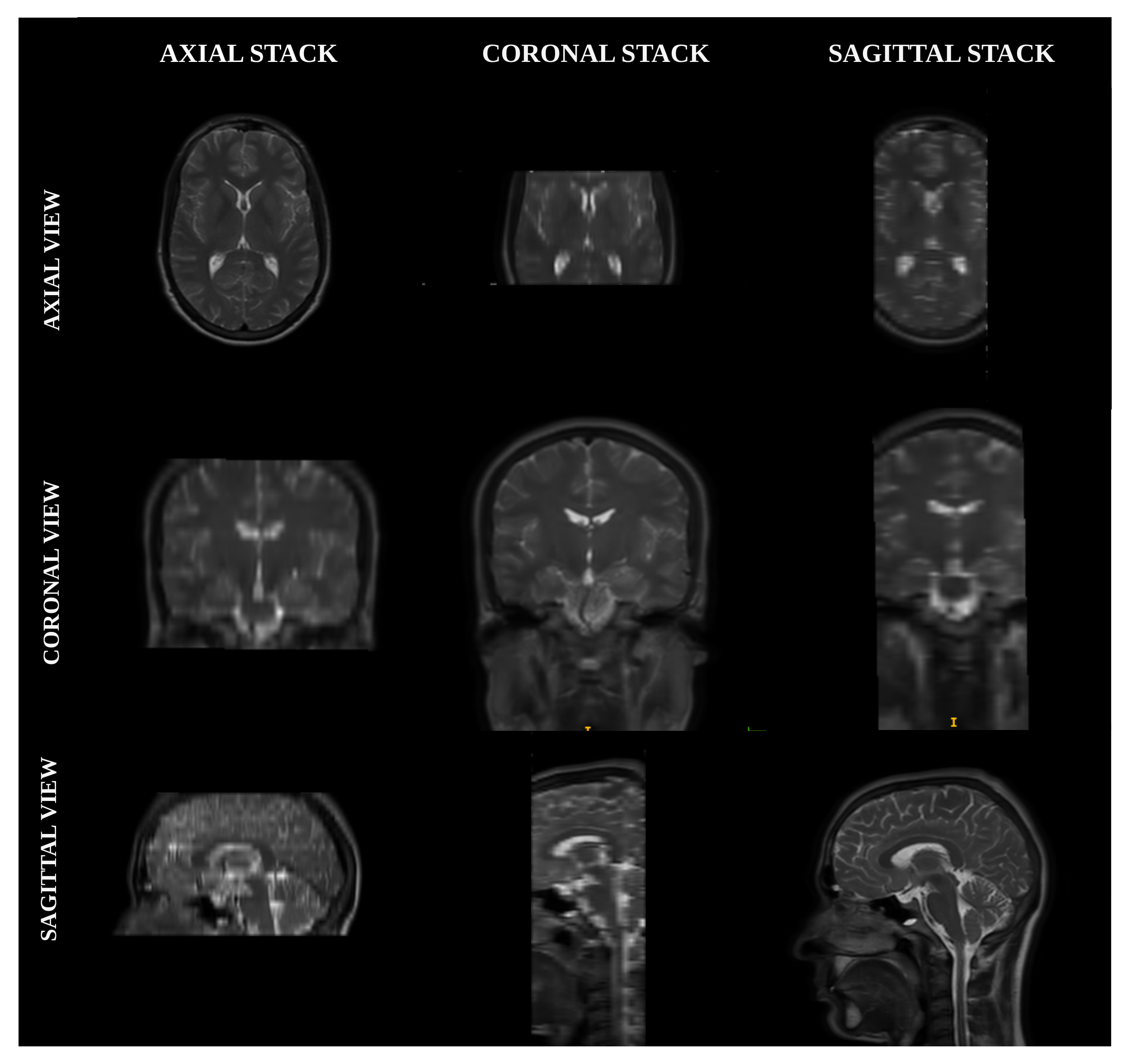}
    \caption{\textbf{Partial-FOV in real multi-orientation clinical acquisitions.}  Different anisotropic stacks may cover different anatomical extents.}
    \label{fig:partial_fov}
\end{figure}

Real multi-orientation clinical acquisitions do not necessarily cover the same anatomical extent. As shown in \Cref{fig:partial_fov}, stacks acquired from the same subject can terminate at different superior-inferior or anterior-posterior locations depending on their orientation and clinical prescription.

\subsection{Reconstruction at multiple resolutions}
\label{appendix:multi_res}

\begin{figure}[pos=htbp]
    \centering
    \includegraphics[width=0.6\textwidth]{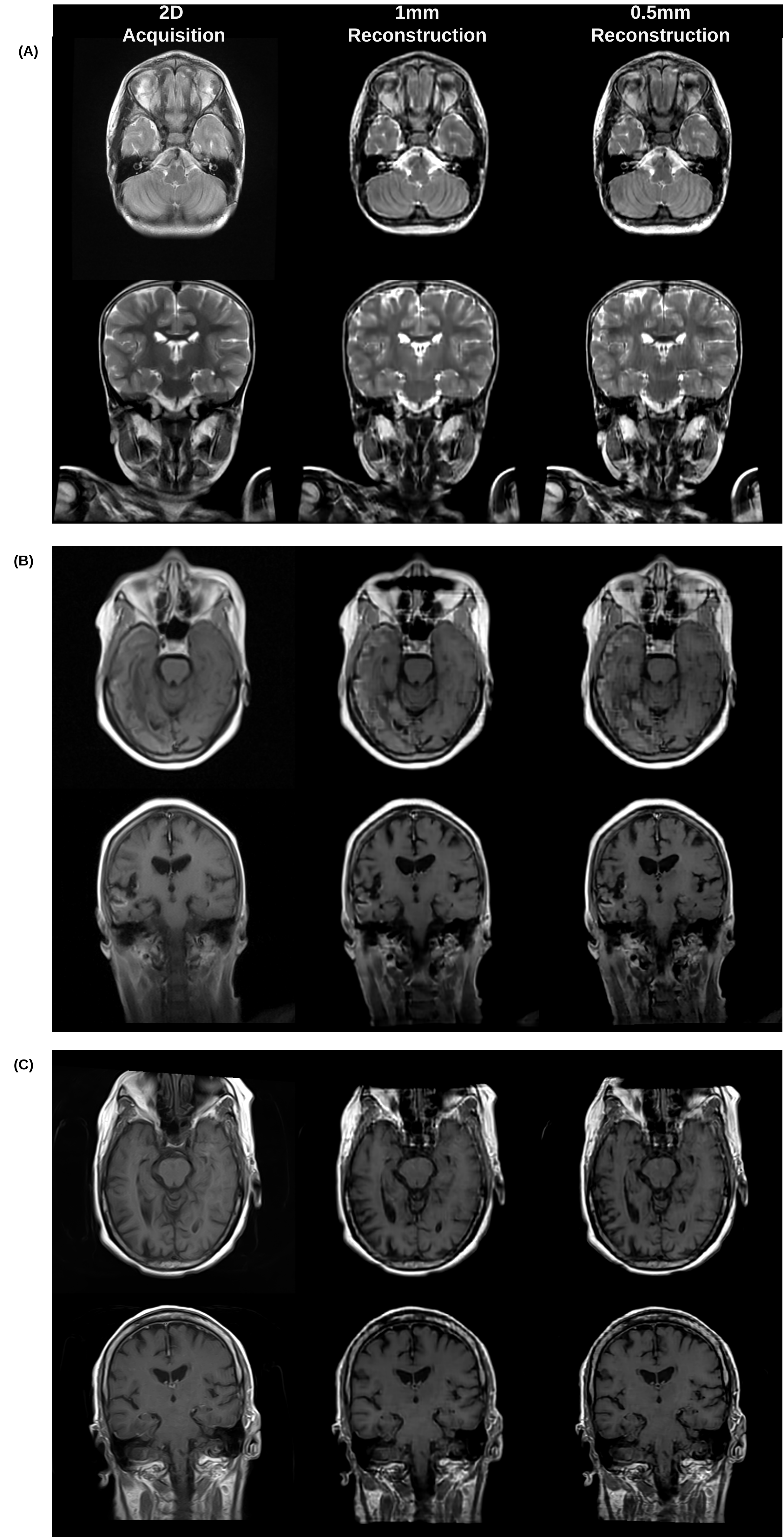}
    \caption{\textbf{MS-HVED reconstruction on different output grids.}
Representative cases reconstructed at $1\,\mathrm{mm}$ and $0.5\,\mathrm{mm}$ isotropic resolution using the same model trained with $1\,\mathrm{mm}$ targets. The left column shows the original 2D acquisitions at their native in-plane resolution. The $0.5\,\mathrm{mm}$ reconstructions were generated using sliding-window inference with $128\times128\times64$ patches and $50\%$ overlap.
\textbf{(A)} $1.5$\,T acquisition with $0.5$\,mm native in-plane resolution and $6.5$\,mm through-plane spacing, reconstruction grids were $262\times290\times286$ at $1\,\mathrm{mm}$ and $521\times577\times569$ at $0.5\,\mathrm{mm}$.
\textbf{(B)} $0.3$\,T acquisition with $0.5$--$1.0$\,mm native in-plane resolution and $7.0$--$8.5$\,mm through-plane spacing, reconstruction grids were $250\times249\times250$ and $497\times496\times497$, respectively.
\textbf{(C)} $1.5$\,T acquisition with $0.4$\,mm native in-plane resolution and $6$\,mm through-plane spacing, reconstruction grids were $292\times299\times243$ and $582\times595\times483$, respectively.}
\label{fig:multi_res}
\end{figure}

As a fully convolutional architecture, MS-HVED is not constrained to the spatial matrix size used during training and can therefore be evaluated on reconstruction grids with different resolutions. Although all models were trained using $1\,\mathrm{mm}$ isotropic targets, we additionally performed inference on a $0.5\,\mathrm{mm}$ isotropic grid without retraining. For the larger output volumes, inference was performed using overlapping $128\times128\times64$ patches with $50\%$ overlap.

\Cref{fig:multi_res} shows three examples spanning heterogeneous clinical acquisitions, including field strengths of $0.3$ and $1.5$\,T, native in-plane resolutions of $0.4$--$1.0$\,mm, and through-plane spacings of $6.0$--$8.5$\,mm. In all cases, MS-HVED can be evaluated directly on the denser $0.5\,\mathrm{mm}$ grid. The $0.5\,\mathrm{mm}$ grids contain approximately $7.9\times$ as many voxels as their $1\,\mathrm{mm}$ counterparts. These examples demonstrate the flexibility of MS-HVED to produce reconstructions on user-defined grids across acquisitions with substantially different spatial resolutions and field strengths. We emphasise, however, that the $0.5\,\mathrm{mm}$ outputs represent inference on a denser reconstruction grid not validation of true $0.5\,\mathrm{mm}$ spatial resolution.

\end{document}